\documentclass[journal]{IEEEtran}

\usepackage{tabularx}

\usepackage{comment}
\usepackage{graphicx}
\usepackage{amsmath}
\usepackage{amssymb}
\usepackage{comment}
\usepackage{multirow}
\usepackage{hyperref}
\hypersetup{hidelinks}

\usepackage{booktabs}
\usepackage{array, caption, threeparttable}
\usepackage{caption}
\usepackage{algorithm}
\usepackage{listings}
\usepackage{color}
\usepackage[dvipsnames, table]{xcolor}
\usepackage{mathtools}

\usepackage{todonotes}
\usepackage{gensymb}

\usepackage{makecell}

\usepackage{xspace}
\definecolor{baselinecolor}{gray}{0.8}

\usepackage{gensymb}
\usepackage{pifont}

\usepackage{amssymb}
\makeatletter

\newcommand{\Rmnum}[1]{\expandafter\@slowromancap\romannumeral #1@}

\usepackage{bbm}

\usepackage{subcaption}
\usepackage{algorithm}
\usepackage{algpseudocode}
\usepackage[table]{xcolor}

\definecolor{b}{rgb}{0,0,0}

\begin{document}

\title{UniEmo: Unifying Emotional Understanding and
Generation with Learnable Expert Queries}

\author{Yijie Zhu,
        Lingsen Zhang,
        Zitong Yu$^\dagger$,~\IEEEmembership{Senior Member,~IEEE},
        Rui Shao$^\dagger$,~\IEEEmembership{Member,~IEEE},
        Tao Tan,~\IEEEmembership{Member,~IEEE},
        Liqiang Nie~\IEEEmembership{Senior Member,~IEEE}
        \\

\thanks{Manuscript received July, 2025. This work was supported by National Natural Science Foundation of China (Grant No. 62306061 and 62576076), Guangdong Basic and Applied Basic Research Foundation (Grant No. 2023A1515140037), CCF-Tencent Rhino-Bird Open Research Fund, SongShan Lake HPC Center (SSL-HPC) in Great Bay University, National Natural Science Foundation of China (Grant No. 62306090), Natural Science Foundation of Guangdong Province of China (Grant No. 2024A1515010147)  and Natural Science Foundation of Shenzhen City of China (Grant No. JCYJ20250604145700001) and Beijing Natural Science Foundation (L254018).
$^\dagger$ Corresponding author: Zitong Yu (email: yuzitong@gbu.edu.cn) and Rui Shao (email: shaorui@hit.edu.cn).}

\thanks{Yijie Zhu is with Harbin Institute of Technology, Shenzhen, Shenzhen 518055, China, and Great Bay University, Dongguan 523000, China.}

\thanks{Lingsen Zhang, and Liqiang Nie are with Harbin Institute of Technology, Shenzhen, Shenzhen 518055, China.}

\thanks{Zitong Yu is with School of Computing and Information Technology, Great Bay University, Dongguan,523000, China, and also with
Dongguan Key Laboratory for Intelligence and Information Technology.}

\thanks{Rui Shao is with Harbin Institute of Technology, Shenzhen, Shenzhen 518055, China, and also with
Shenzhen Loop Area Institute.}

\thanks{Tao Tan is with Macao Polytechnic University, Macao 999078, China.}

}


\maketitle

\begin{abstract}
Emotional understanding and generation are often treated as separate tasks, yet they are inherently complementary and can mutually enhance each other. In this paper, we propose the \textbf{UniEmo}, a unified framework that seamlessly integrates these two tasks. The key challenge lies in the abstract nature of emotions, necessitating the extraction of visual representations beneficial for both tasks. To address this, we propose a \textbf{hierarchical emotional understanding chain with learnable expert queries} that progressively extracts multi-scale emotional features, thereby serving as a foundational step for unification. Simultaneously, we fuse these expert queries and emotional representations to guide the diffusion model in generating emotion-evoking images. To enhance the diversity and fidelity of the generated emotional images, we further introduce the emotional correlation coefficient and emotional condition loss into the fusion process. This step facilitates \textbf{fusion and alignment for emotional generation guided by the understanding.} In turn, we demonstrate that joint training allows the generation component to provide implicit feedback to the understanding part. Furthermore, we propose a novel data filtering algorithm to select high-quality and diverse emotional images generated by the well-trained model, which explicitly feedback into the understanding part. Together, these \textbf{generation-driven dual feedback processes enhance the model's understanding capacity.} Extensive experiments show that UniEmo significantly outperforms state-of-the-art methods in both emotional understanding and generation tasks. The code for the proposed method is available at \url{https://github.com/JiuTian-VL/UniEmo} 
\end{abstract}


\IEEEpeerreviewmaketitle

\vspace{-1em}
\section{Introduction}
\begin{figure*}[!h]
    
    \centering
    \includegraphics[width=1.0\textwidth]{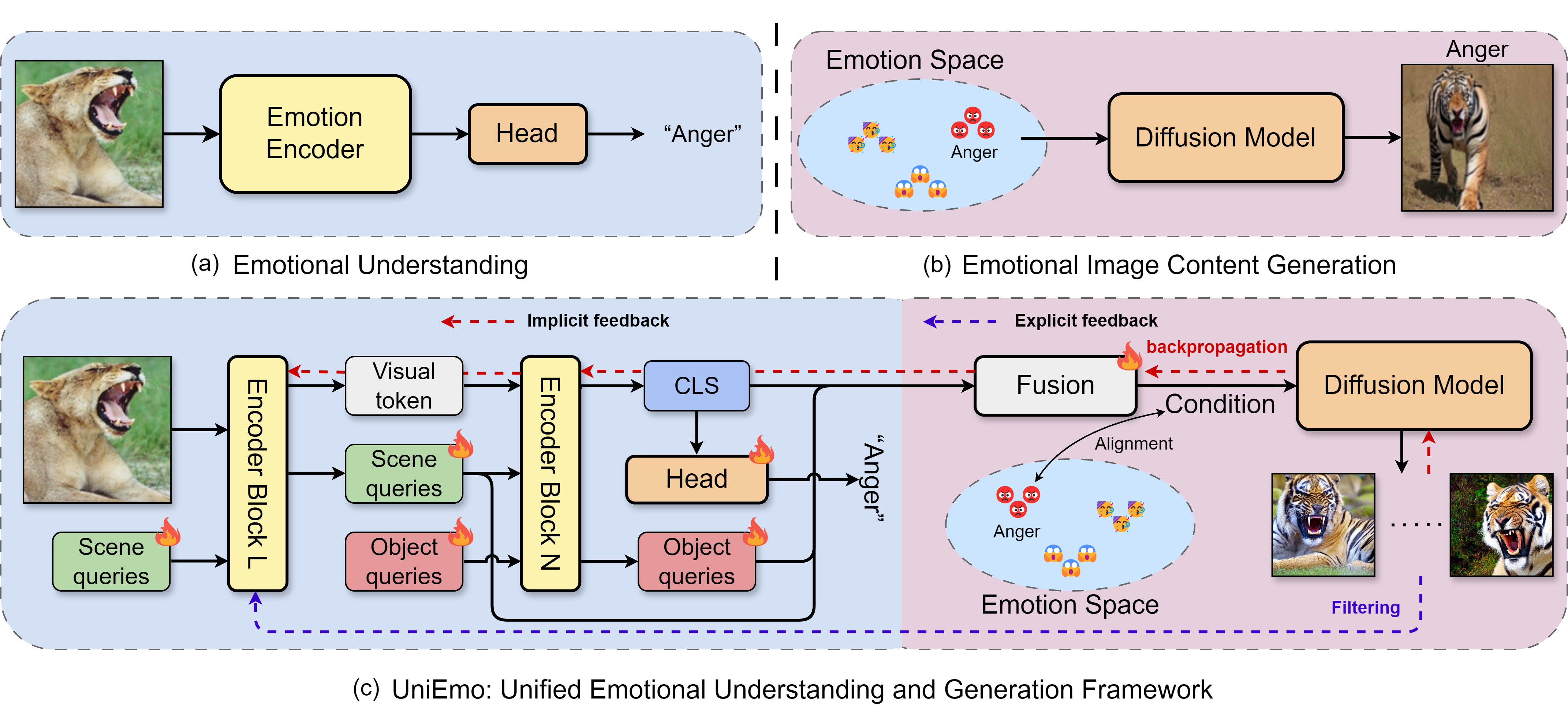}
    \vspace{-15pt}
    \caption{The pipelines for existing methods focused on Emotional Understanding and Emotional Image Content Generation are shown in \textbf{(a)} and \textbf{(b)}, respectively. In contrast, \textbf{(c)} illustrates our unified framework, \textbf{UniEmo}, which integrates both tasks. This framework comprises \textbf{three key steps}: the blue section indicates the first step, the pink section corresponds to the second step, and the dashed lines represent the third step. \textcolor{b}{CLS denotes the class token used for classification, and the flame icon marks trainable modules. Detailed explanations of the expert queries (scene and object queries) are provided in Sec.~\ref{sec: 3-1}.}}
    \label{fig:introduction}
    \vspace{-10pt}
\end{figure*}
Emotional understanding has gained significant attention due to its crucial role in various domains such as human-computer interaction~\cite{cheng2017video, cheng2017video2shop, lyu2025puma,zhu2026delta, shao2019multi, li2025cogvla, zhu2026H-GAR, li2025semanticvla}, mental health~\cite{hutchison2017emotion} and entertainment~\cite{yang2021stimuli}. Existing approaches~\cite{zhang2011analyzing, chen2015learning,zhu2017dependencyEA,zhang2020exploring,reddy2021emotion,yang2021solver} treat visual emotion understanding as a conventional classification task, as illustrated in Fig.~\hyperref[fig:introduction]{1(a)}. Typically, a visual encoder extracts features, which are then mapped to target emotion using a classification head.

    Beyond emotional understanding, Emotional Image Content Generation (EICG) has emerged as a key research direction~\cite{yang2024emogen, zhu2025emosym, yuan2025coemogen}.  This generative task is particularly valuable in digital media, virtual environments, and marketing~\cite{yadollahi2017current, hsieh2015conceptualizing, xie2023learning}.  Current works commonly create a latent emotion space in which similar emotional representations are grouped closely, while dissimilar ones are separated. To generate an image that embodies a specific emotion, the corresponding emotional feature extracted from this latent space is used as a conditioning input to guide the diffusion model~\cite{ho2020denoising, rombach2022high, dhariwal2021diffusion, zhang2023adding, ruiz2023dreambooth}, as shown in Fig.~\hyperref[fig:introduction]{1(b)}.

    Although emotional understanding and generation are generally treated as separate tasks, they are inherently complementary~\cite{xie2024show,zhou2024transfusion,wu2024vila, wang2024emu3}. Enhancing emotional generation depends on a deeper emotional understanding~\cite{yang2024emogen}, as a better comprehension of emotional cues leads to more expressive image generation. Meanwhile, emotional generation can enhance emotional understanding by expanding limited emotion datasets with diverse, synthesized data~\cite{li2024id}. Additionally, it provides generative feedback that refines distinctive emotional features, further enhancing emotional understanding~\cite{wang2024diffusion}.

    Motivated by this synergy, we aim to design a unified framework that integrates them, enabling mutual reinforcement between the two tasks. To facilitate this integration, it is essential to identify a pivotal entry point. The core pipelines in existing methods, as depicted in Fig.~\hyperref[fig:introduction]{1(a)} and~\hyperref[fig:introduction]{(b)}, reveal that extracting emotion-relevant and semantically rich visual representations is essential for both tasks, which is the key to unify them. Specifically, a more accurate understanding of the emotional content in images requires richer emotional representations. Meanwhile, these representations can reinforce a more nuanced emotion space, which helps produce higher-quality emotional images. To extract such richer emotional representations, we need a more fine-grained emotional understanding capability.  Building upon this, instead of focusing solely on advanced feature extractors~\cite{zhang2020exploring,reddy2021emotion,yang2021solver}, we propose decomposing the abstract emotional understanding process into a hierarchical visual chain. It starts with scene-level information, followed by object identification, both of which are crucial for emotion recognition~\cite{yang2021solver, bar2004visual, frijda2009emotion}, and then leverages these multi-scale prior features to infer emotions. 

    Building on the aforementioned ideas, we introduce two sets of learnable expert queries to extract hierarchical rich representations, thereby constructing a unified emotion understanding and generation framework based on them. Specifically, there are three key steps to build such a framework: \textbf{1)\textit{ Hierarchical Emotional Understanding Chain with Expert Queries.}}  The two sets of expert queries serve different purposes: one set focuses on extracting scene-level features, while the other targets object-level features. As illustrated by the blue section in Fig.~\hyperref[fig:introduction]{1(c)}, they are progressively integrated into the network to extract hierarchically rich emotional representations. \textbf{2)\textit{ Understanding-Guided Fusion and Alignment for Emotional Generation.}} To condition the diffusion model for generating emotionally distinct and diverse images, we fuse these expert queries with emotional representations, as depicted by the pink section in Fig.~\hyperref[fig:introduction]{1(c)}. However, simple fusion risks weakening emotional signals and misaligning representations with the target emotion space. To address these issues, we introduce an emotional correlation coefficient to quantify the emotional relevance of expert queries and an emotional condition loss to align outputs with the target emotion space using contrastive learning. \textbf{3)\textit{ Generation-Driven Dual Feedback for Enhanced Emotional Understanding.}} We demonstrate that joint training enables the generation component to provide implicit feedback that optimizes discriminative emotional representations and enhances understanding~\cite{wang2024diffusion}. Additionally, we propose a novel data filtering algorithm to select high-quality and diverse emotional images generated by the well-trained model. By enriching the limited emotion dataset with such synthesized examples, this process provides explicit feedback to further strengthen the model's understanding ability. The two types of feedback are represented by dashed lines in Fig.~\hyperref[fig:introduction]{1(c)}. In summary, our contributions are as follows:
    \vspace{-10pt}
    \begin{itemize}
    \item We introduce \textbf{UniEmo}, a novel framework that integrates emotional understanding and generation through \textbf{a hierarchical sequence of learnable expert queries}. These queries enable the progressive extraction of rich emotional features, forming the foundation for unification.
    \item We propose the \textbf{emotional correlation coefficient} for efficient fusion and design an \textbf{emotional condition loss} to align the fused features with the target emotion space, enhancing the quality of generated emotional images.
    \item We demonstrate that through joint training, the generation component provides \textbf{implicit feedback} to the understanding. Additionally, we propose a novel data filtering algorithm that \textbf{explicitly feeds back} into the understanding component, further enhancing performance.
    \item We conduct extensive experiments to validate UniEmo, demonstrating its superior performance in both emotional understanding and generation tasks.
    \end{itemize}
\vspace{-10pt}
\section{Related Work}
\subsection{Visual Emotion Understanding}
Visual emotion understanding has attracted increasing attention in recent years~\cite{zhao2021affective}. Early approaches primarily relied on low-level visual features such as color, texture, and shape to infer emotional states~\cite{lee2011fuzzy,borth2013large,jana2010affective}. With the advent of deep learning, convolutional neural networks (CNNs) have enabled the automatic learning of hierarchical features, ranging from edges and textures to complex shapes and patterns~\cite{xu2022mdan,rao2020learning,yang2021stimuli}, thereby capturing nuanced emotional expressions beyond pixel-level details. \textcolor{b}{Moreover, some works have explored robustness-oriented designs to cope with real-world imaging degradations.}
\textcolor{b}{For instance, LRDif~\cite{wang2024lrdif} leverages diffusion-based distribution mapping and transformer-based dependency modeling to improve FER under under-display camera distortions.}
Despite these advancements, many recent studies~\cite{zhang2020object,borth2013sentibank,chen2015learning,pan2023progressive} continue to focus primarily on enhancing visual feature extraction through increasingly complex neural architectures. 
Importantly, a growing body of evidence highlights the crucial role of high-level semantic concepts, such as scene context, salient objects, and their interactions, in shaping emotional perception~\cite{yang2021solver, bar2004visual, frijda2009emotion}. 

To this end, recent methods have started to incorporate structured semantic priors and auxiliary information. 
\textcolor{b}{SoVTP~\cite{wang2025visual} integrates spatial annotations, facial action units, and contextual cues into unified prompts for holistic emotion recognition.}
However, a systematic framework for modeling emotional understanding through multi-scale semantic reasoning is still lacking. To bridge this gap, our approach explicitly integrates scene-level and object-level representations via hierarchical expert queries, which have been validated to offer complementary and discriminative cues for robust emotion recognition.
\vspace{-10pt}
\subsection{Visual Emotion Generation}
Visual emotion generation aims to synthesize visual content that evokes specific emotions by leveraging advanced generative models and affective computing techniques. Recently, diffusion models have achieved remarkable progress, giving rise to powerful frameworks such as GLIDE~\cite{nichol2021glide}, DALLE2~\cite{ramesh2022hierarchical}, eDiff-I~\cite{balaji2023ediffi}, and Imagen~\cite{saharia2022photorealistic}. These models excel at generating concrete concepts~\cite{dhariwal2021diffusion,zhang2023adding,liao2022text} and personalized imagery~\cite{gal2022image,ruiz2023dreambooth,kumari2023multi}, but often struggle with producing abstract, emotion-evoking visuals.

To address this challenge, EmoGen~\cite{yang2024emogen} introduces an emotion space aligned with CLIP semantics~\cite{radford2021learning,rombach2022high,wei2023elite} to enhance the emotional expressiveness of generated content. Similarly, EmoEdit~\cite{yang2024emoedit} employs a vision-language model to identify emotion-relevant semantic factors and guides the generative model to modify images accordingly. EmotionPrompt\cite{liemotionprompt} formulates emotional intent as learnable soft prompts, enabling emotion-conditioned generation by adapting the internal attention flow of diffusion models. EmotiCrafter\cite{he2025emoticrafter} proposes a framework that models continuous affective dimensions—such as valence and arousal—and integrates them into the diffusion process to generate emotionally vivid and nuanced content. 
\vspace{-10pt}
\subsection{Query-Based Methods}
Query-based mechanisms~\cite{zhang2025falcon, ye2024mplug, bai2023qwen, li2025lion, chen2024lion} have gained increasing attention in visual representation learning for their ability to decouple task-specific information extraction from raw image features. DETR~\cite{carion2020end} introduces object queries within a Transformer architecture to detect and localize physical entities, optimized through bipartite matching. Slot Attention~\cite{locatello2006object} formulates unsupervised scene decomposition as an iterative querying process, where each slot learns to attend to a distinct object-like component. Subsequent works such as ViTDet~\cite{li2022exploring} and DINO~\cite{zhang2022dino} extend this paradigm for object detection and dense prediction using query-based Transformer designs. In the vision-language domain, approaches like MaskCLIP~\cite{zhou2022extract} and DenseCLIP~\cite{rao2022denseclip} employ language-driven prompts or queries to guide fine-grained semantic alignment between textual concepts and visual regions.

While these methods are effective for object detection and semantic alignment, they largely neglect abstract emotional semantics. In contrast, our approach introduces hierarchical expert queries—scene-level and object-level—specifically designed for emotional understanding and generation. This design is motivated by prior evidence highlighting the critical role of scenes and objects in emotion perception.
Moreover, UniEmo employs a closed-loop framework where expert queries facilitate both recognition and generation, enabling synergistic learning and feedback across tasks.
\section{METHODOLOGY}
\label{sec:UniEmo}
\begin{figure*}
    \centering
    \includegraphics[width=0.98\textwidth]{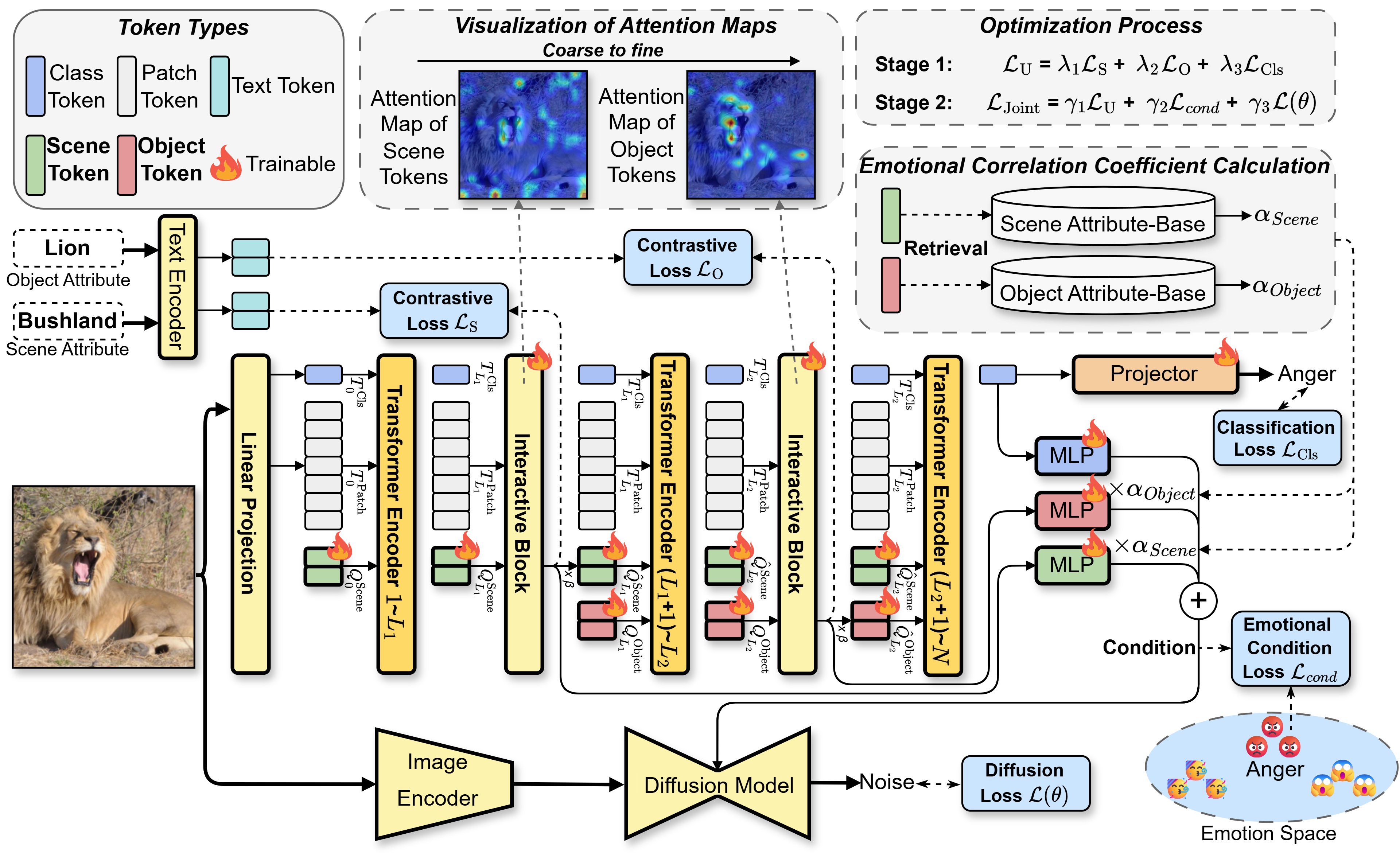}
    \vspace*{-2mm} 
    \caption{Overview of our \textbf{UniEmo} framework, which leverages \textbf{two sets of learnable expert queries} \textbf{(scene tokens and object tokens)} to capture hierarchical emotional representations (step 1 mentioned in the introduction). These tokens and emotional representations are fused with \textbf{emotional correlation coefficients} $\alpha_{\text{Scene}}$ and $\alpha_{\text{Object}}$, supervised by the \textbf{emotional condition loss} $L_{\text{cond}}$ to guide the diffusion model in emotion-driven generation (step 2). Similar to Fig.~\hyperref[fig:introduction]{1(c)}, the generation component provides \textbf{dual feedback} to the understanding module, though this is not explicitly shown in this figure (step 3). The top section of the figure shows the attention maps for the expert queries.}
    \label{fig:overall_method}
    \vspace*{-6mm}
\end{figure*}
In this section, we provide a detailed explanation of UniEmo. As shown in Fig.~\hyperref[fig:introduction]{1}, the model operates within a closed-loop framework built around three key components: 1) a hierarchical emotional understanding chain with expert queries (blue section), 2) efficient fusion and alignment with these expert queries for emotional generation (pink section), and 3) dual feedback from the generation to the understanding (dashed lines). We will elaborate on these components sequentially in Secs.~\ref {sec: 3-1}–\ref{sec: 3-3}.
\vspace{-10pt}
\subsection{Hierarchical Emotional Understanding Chain with Expert Queries}
\label{sec: 3-1}
 For an input image, we adopt the ViT~\cite{alexey2020image} by dividing it into non-overlapping patches, linearly projecting them into patch tokens, and prepending a class token to the sequence. For convenience, in an $N$-layer ViT model, the class and patch tokens output from the $m$-th layer are denoted as  \(T_{m}^{\text{Cls}}\) and \(T_{m}^{\text{Patch}}\) (\( 0 \leq m \leq N \)), respectively. Notably, \(T_{0}^{\text{Cls}}\) and \(T_{0}^{\text{Patch}}\) represent the input tokens before the first encoding layer.
 
\vspace{2pt}
 \noindent\textbf{Multi-stage Expert Queries.}
 Instead of directly feeding the tokens into the forward pass to infer abstract emotions, we introduce two sets of learnable expert queries that decompose this process into a hierarchical visual chain. As shown in Fig.~\ref{fig:overall_method}, the two expert queries, denoted as \(Q_{m}^{\text{Scene}}\) and \(Q_{m}^{\text{Object}}\), are a series of learnable tokens designed to extract scene-level and object-level features, respectively. \textcolor{b}{These query embeddings are model parameters shared by all inputs. They are randomly initialized and optimized end-to-end. We divide the \(N\)-layer Transformer architecture into three hierarchical stages: layers 1 to \(L_1\), \(L_1+1\) to \(L_2\), and \(L_2+1\) to \(N\). We inject them hierarchically across the Transformer: scene queries are introduced at the input to capture global context, and object queries are subsequently added to emphasize object-centric cues. To make each query type specialize, we apply level-specific supervision to encourage attention to the intended granularity, as detailed below. With this supervision, the model forms a hierarchical understanding chain, focusing on scene cues in layers \(1\!\sim\!L_1\), object cues in layers \(L_1{+}1\!\sim\!L_2\), and their joint integration in layers \(L_2{+}1\!\sim\!N\).}

 Beginning with the first layer, we initially concatenate the \(T_{0}^{\text{Cls}}\), \(T_{0}^{\text{Patch}}\) and \(Q_{0}^{\text{Scene}}\), and then pass them through the encoder layers from 1 to \(L_1\), represented as:
 \setlength{\abovedisplayskip}{3pt}
        \begin{align}
            \left[ T_{L_1}^{\text{Cls}},\! T_{L_1}^{\text{Patch}},\! Q_{L_1}^{\text{Scene}} \right] = \text{E}_{1:L_1} \left( \left[ T_{0}^{\text{Cls}},\! T_{0}^{\text{Patch}},\! Q_{0}^{\text{Scene}} \right] \right),
        \end{align}  
where \( \mathrm{E}_{1:L_1} \) denotes the Transformer encoder block from layer 1 to \( L_1 \), \( \left[\cdot, \cdot \right] \) represents concatenation operation along the sequence dimension. The previous steps established an initial correlation between the expert query and image tokens. To strengthen this interaction, we introduce an Interactive Block, which facilitates the extraction of more detailed scene features from the image tokens. Specifically, this block functions as a compact transformer-style network with two layers, each comprising a multi-head attention mechanism, represented as \( {A} \), followed by a feed-forward network, denoted as \( {F} \). We input  \(Q_{L_1}^{\text{Scene}}\) and \(T_{L_1}^{\text{Patch}}\) into the Interactive Block, with the computation process defined as:
   \begin{equation}
        \begin{aligned}
            \bar{Q}_{L_1}^{\text{Scene}} &= Q_{L_1}^{\text{Scene}} + {A}(Q_{L_1}^{\text{Scene}}, T_{L_1}^{\text{Patch}}), \\
            \tilde{Q}_{L_1}^{\text{Scene}} &= \bar{Q}_{L_1}^{\text{Scene}} + {F}(\bar{Q}_{L_1}^{\text{Scene}}).
        \end{aligned}
        \end{equation}
Subsequently, we enhance the original \(Q_{L_1}^{\text{Scene}}\) using the refined \( \tilde{Q}_{L_1}^{\text{Scene}} \), which is derived from the output of the Interactive Block, following the procedure outlined below:
 \begin{equation}
            \hat{\mathit{Q}}_{L_1}^{\text{Scene}} = \mathit{Q}_{L_1}^{\text{Scene}} + \beta \tilde{\mathit{Q}}_{L_1}^{\text{Scene}},
        \end{equation}
where $\beta$ represents the learnable modulation parameter, enabling the model to effectively balance the contributions of both the primary and refined scene-level expert query.    

To guide the expert query in extracting scene-level representations from emotional images, we use a contrastive loss to enforce semantic supervision and alignment. As illustrated in Fig.~\ref{fig:overall_method}, for an example of a lion image, we use a CLIP~\cite{radford2021learning} text transformer to encode the corresponding scene attribute \textquotedblleft bushland\textquotedblright ~as the supervisory signal. Formally, given a batch of \( K \) query-attribute pairs \( \{(\hat{\mathit{Q}}_{L_1,i}^{\text{Scene}}, z_{i}^{\text{Scene}})\}_{i=1}^{K} \), where \( z_{i}^{\text{Scene}} \) represents the text embedding of corresponding scene attribute for the \( i \)-th pair, the contrastive loss \( \mathcal{L}_{\text{S}} \) is thus defined as:
        \begin{align}
            \mathcal{L}_{\text{S}} &= - \frac{1}{K} \sum_{i=1}^{K}  \log \frac{\exp \left( \text{sim} \left( \hat{\mathit{Q}}_{L_1,i}^{\text{Scene}}, z_{i}^{\text{Scene}} \right) / \tau \right)}{\sum_{j=1}^{K} \exp \left( \text{sim} \left( \hat{\mathit{Q}}_{L_1,i}^{\text{Scene}}, z_{j}^{\text{Scene}} \right) / \tau \right)},
        \end{align} 
where \( \text{sim}(\cdot, \cdot) \) denotes the dot product similarity, and \( \tau \) is a temperature hyperparameter, which is set to 0.07.
Following the procedures outlined above, we enable the model to form an initial understanding of the global context of the entire image. Building on this prior, we introduce another expert query dedicated to extracting object-level features. Specifically, we concatenate the \(T_{L_1}^{\text{Cls}}\), \(T_{L_1}^{\text{Patch}}\), \(\hat{\mathit{Q}}_{L_1}^{\text{Scene}}\) and \(Q_{L_1}^{\text{Object}}\) at the input of layer \(L_1+1\). Similar to the previous process, they are then passed through Transformer encoders from \(L_1+1\) to \(L_2\), after which \(Q_{L_2}^{\text{Object}}\) and \(T_{L_2}^{\text{Patch}}\) are input into the Interactive Block. Object attributes encoded by the text encoder guide this process,  producing a contrastive loss \( \mathcal{L}_{\text{O}} \). Finally, the two sets of expert queries work together in the encoding layers from \(L_2+1\) to \(N\). Thus, a hierarchical visual chain emerges, enabling a novel emotional understanding process that identifies abstract emotions and extracts multi-scale features for unification.
Notably, scene and object attributes are used solely during training to supervise expert query learning. Once trained, the model requires only a single image at inference time to perform emotion prediction.
\vspace{-10pt}
\subsection{Understanding-Guided Fusion and Alignment for Emotional Generation}
\label{sec: 3-2}
\vspace{-5pt}
From the emotional understanding chain, we observe that the class token before the classification head encodes distinct emotional representations, while multi-stage expert queries capture diverse semantic features. By complementing these two components to form a clear and diverse emotion space~\cite{yang2024emogen}, we can better guide the diffusion model to generate emotionally distinct and diverse images. However, multi-stage expert queries span a hierarchical spectrum of scene and object characteristics whose emotional valence can vary significantly. Some elements exhibit strong emotional resonance (e.g., Ferris wheels with amusement or flames with anger), while others are emotionally neutral (e.g., trees and grass). To address this variability, we propose an emotional correlation coefficient to quantitatively evaluate the relationship between expert queries and emotions. This coefficient is a pivotal metric for assessing the significance of these queries in the fusion process.
\begin{figure}
    \centering
    \includegraphics[width=\linewidth]{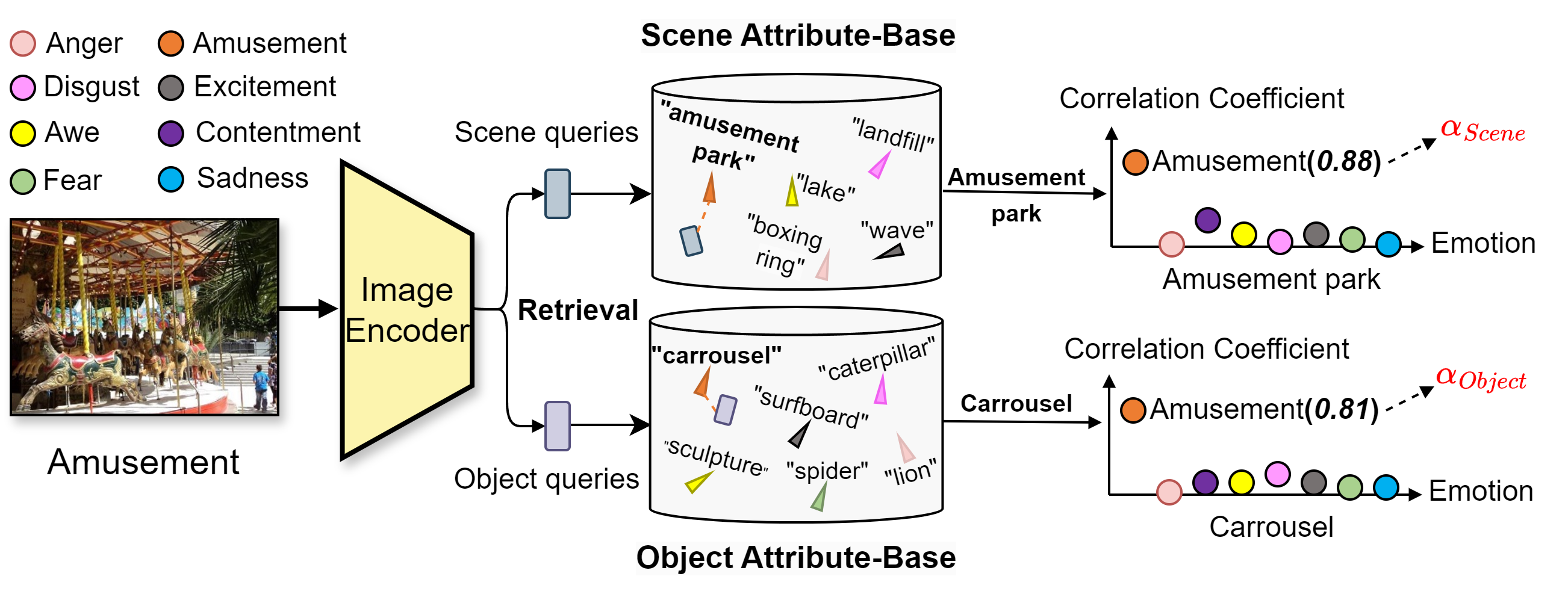}
    \caption{The overall pipeline for calculating the emotional correlation coefficient of expert queries.}
    \vspace{-15pt}
    \label{fig:coefficient}
    \vspace{-8pt}
\end{figure}

\vspace{2pt}
\noindent\textbf{Emotional Correlation Coefficient.} 
As shown in Fig.~\ref{fig:coefficient}, 
we consolidate all scene and object attributes from the dataset to construct the corresponding attribute base. 
For each type of expert query, we identify the most semantically relevant attributes using cosine similarity. Subsequently, all images possessing this attribute are input into a pre-trained emotion classifier, which outputs the log probabilities associated with this attribute. To quantify the relevance of an attribute \(j\) to a specific emotion
\(e\), we calculate the emotional correlation coefficient as follows:
  \setlength{\abovedisplayskip}{3pt}
        \begin{align}
        \alpha_{j,e}  = \frac{1}{M_j} \sum_{k=1}^{M_j} \ell(I_k, e),
    \label{eq:correlation}
    \vspace{-5pt}
    \end{align}
    where \( I_k \) represents an input image, \( \ell(I_k, e) \) denotes its log probability of being classified under \( e \), and \( M_j \) is the number of images that contain attribute \( j \).    

Subsequently, we leverage this correlation coefficient to effectively integrate the features of the previously mentioned two expert queries and the class token, serving as conditional guidance for the diffusion model. In detail, as depicted in Fig.~\ref{fig:overall_method}, we employ separate Multilayer Perceptrons (MLPs) \( \phi_1 \), \( \phi_2 \), and \( \phi_3 \) for each of these features, mapping them individually into the input space of the diffusion model. Emotional correlation coefficients \( \alpha_{\text{Scene}} \) and \( \alpha_{\text{Object}} \), computed for each expert query using Eq.~\ref{eq:correlation}, are applied  as weights during the fusion process, leading to the condition \(c\):
        \begin{equation}
             c = \alpha_{\text{Scene}}\cdot\phi_1(\hat{Q}_{L_1}^{\text{Scene}}) \oplus \alpha_{\text{Object}}\cdot\phi_2(\hat{Q}_{L_2}^{\text{Object}}) \oplus \phi_3(T_N^{\text{Cls}}),
        \end{equation}
where \( \hat{Q}_{L_1}^{\text{Scene}} \), \( \hat{Q}_{L_2}^{\text{Object}} \), and \( T_N^{\text{Cls}} \) are expert queries and class token, and \( \oplus \) is element-wise addition. 

\vspace{2pt}
\noindent\textbf{Emotional Condition Loss.}
 Building on the aforementioned coefficients, to explicitly supervise the fused condition \(c\), ensuring it aligns with the intended emotion space, we propose an emotional condition loss. Specifically, we leverage the principles of contrastive learning, where each emotional image and its true emotional category   \( e^+ \) serve as the positive sample. The selection of negative sample \( e^- \) is guided by the model’s understanding component: if the emotional category is correctly predicted, the second-highest scoring category is chosen as the negative sample; otherwise, the misclassified category is used. We add an additional classification head for the condition \(c\), yielding the emotion distribution \( e^p \),  which is a probability vector over the emotion categories. For the triplet \( (e^+, e^-, e^p) \), the emotional condition loss \( \mathcal{L}_{cond}\) is defined as:
  \setlength{\abovedisplayskip}{3pt}
        \begin{align}
            \mathcal{L}_{cond} = \max\left( 0, d(e^p, e^+) - d(e^p, e^-) + \xi \right),
        \end{align}
 where \( d(\cdot, \cdot) \) represents the cosine distance, and \( \xi \) is a margin scalar that enforces a minimum separation between the positive and negative samples, which is set to 0.1.

 The implicit condition \(c\) is processed by a CLIP text transformer \( \tau \) as done in previous work~\cite{yang2024emogen} to produce textual embeddings that guide the denoising network  \( \epsilon_{\theta} \) in generating emotionally clear and diverse images. As shown in Fig.~\ref{fig:Textinversion}, for a specific emotional text, such as \textquotedblleft fear,\textquotedblright~a special token \(S_{\ast}^{\text{fear}}\) is added to the vocabulary. The previously extracted multi-scale emotional condition feature 
\(c\) is used as the word embedding for the corresponding emotional text during the embedding lookup process. These embeddings are subsequently encoded by a text transformer and fed into the diffusion model.
The training objective \( \mathcal{L}(\theta) \) of this process is described as follows:
         \begin{figure}
            \centering
            \includegraphics[width=1\linewidth]{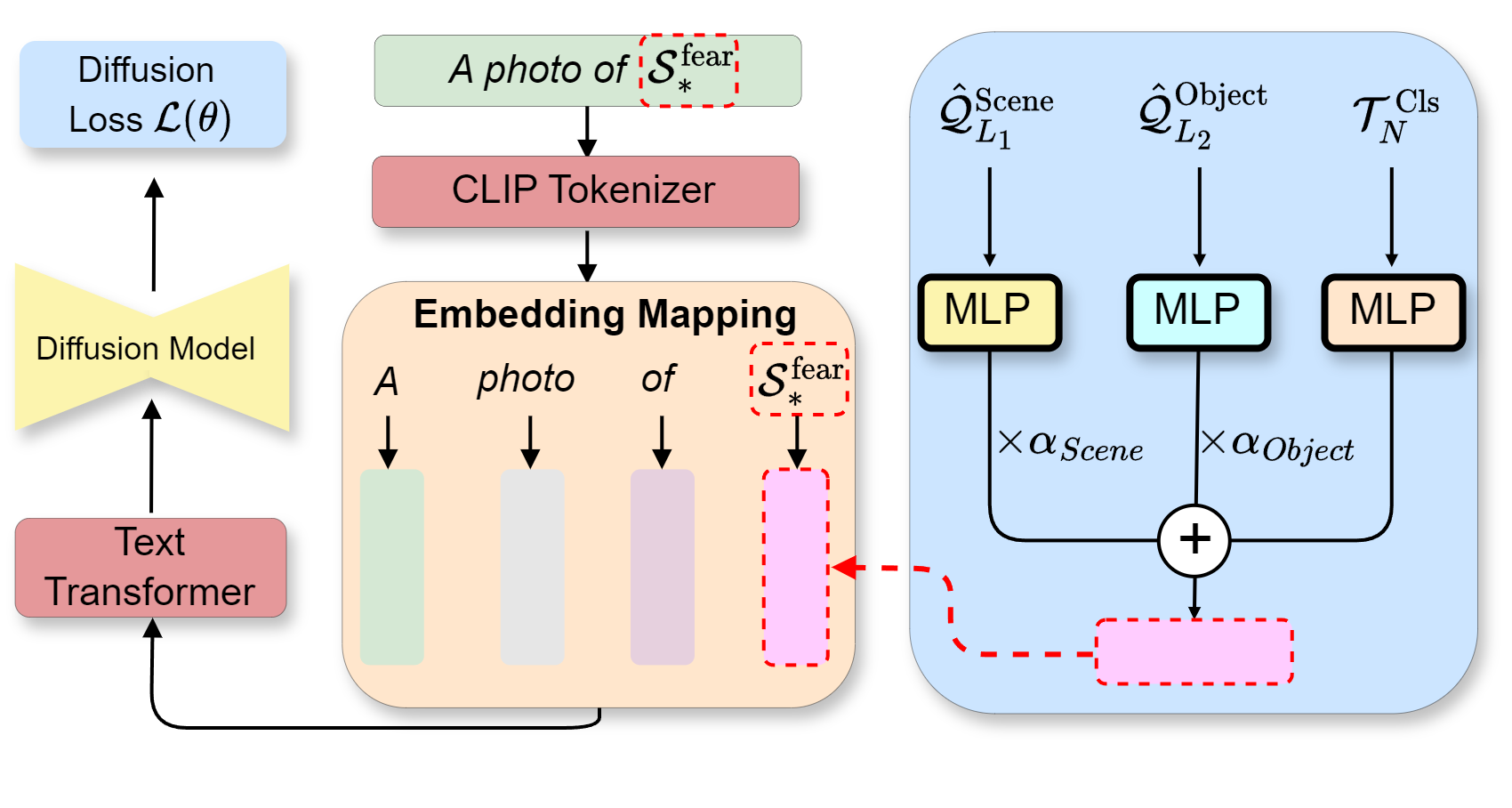}
            \vspace{-15pt}
            \caption{The detailed process of emotional feature fusion and its incorporation as conditional input into the diffusion model.}
            \label{fig:Textinversion}
            \vspace{-20pt}
        \end{figure}
    \begin{equation}
        \begin{aligned}
            \mathbf{x}_t &= \sqrt{\bar{\alpha}_t} \mathbf{x}_0 + \sqrt{1 - \bar{\alpha}_t} \epsilon, \\
            \mathcal{L}(\theta) &= \mathbb{E}_{t, x_0, \epsilon, c} \left[ \left\| \epsilon - \epsilon_{\theta} \left( \mathbf{x}_t, t, \tau(c) \right) \right\|^2 \right],
        \end{aligned}
    \end{equation}
    where \( \mathbf{x}_t \) represents the noisy data at timestep \( t \), combining the original data \( \mathbf{x}_0 \) with Gaussian noise \( \epsilon \). The term \( \bar{\alpha}_t \) is a time-dependent variance schedule. 
    
    During inference, similar to EmoGen~\cite{yang2024emogen}, each emotion cluster is represented by a Gaussian distribution with parameters, i.e., mean and standard deviation. Specifically, upon completion of training, we compute the corresponding mean and variance for the emotional representations \(\hat{Q}_{L_1}^{\text{Scene}}\textbf{}\), \(\hat{Q}_{L_2}^{\text{Object}}\), \(T_N^{\text{Cls}}\) associated with each emotion. During inference, given an emotion label, we sample the corresponding features. These sampled features are then fused using the same integration method as in the training phase and subsequently fed into the diffusion model to generate emotional images.

\vspace{-10pt}

\subsection{Generation-Driven Dual Feedback for Enhanced Emotional Understanding}
\label{sec: 3-3}
  Enhanced understanding improves the quality of generated images. In turn, the generation component provides both implicit and explicit feedback, further refining the model's understanding capability.

\vspace{5pt}
\noindent\textbf{Implicit Feedback.}
 We use a stage-wise training strategy for UniEmo, as shown in Fig.~\ref{fig:overall_method}. In the first stage, we train the understanding component with the following objective:
        \begin{align}
            \mathcal{L}_{\text{U}} = \lambda_1 \mathcal{L}_{\text{S}} + \lambda_2 \mathcal{L}_{\text{O}} + \lambda_3 \mathcal{L}_{\text{Cls}},
        \end{align}
   where \( \mathcal{L}_{\text{S}} \) and \( \mathcal{L}_{\text{O}} \) represent the contrastive losses from the two expert queries discussed in Sec.~\ref{sec: 3-1}, and \( \mathcal{L}_{\text{Cls}} \) denotes the emotional classification loss. \( \lambda_1 \), \( \lambda_2 \), and \( \lambda_3 \) are tuning coefficients, which are set to 0.25, 0.25, and 0.5, respectively. In the second stage, we jointly train the two tasks:
        \begin{align}
            \mathcal{L}_{\text{Joint}} = \gamma_1 \mathcal{L}_{\text{U}} + \gamma_2 \mathcal{L}_{cond} + \gamma_3 \mathcal{L}(\theta),
        \end{align}  
 where the total loss \( \mathcal{L}_{\text{Joint}} \) incorporates \( \mathcal{L}_{\text{U}} \) from the first stage, alongside the emotional condition loss \( \mathcal{L}_{cond} \) and the diffusion loss \( \mathcal{L}(\theta) \). The tuning coefficients \( \gamma_1 \), \( \gamma_2 \), and \( \gamma_3 \) are introduced to balance the contributions of these components during joint training, and are set to 0.3, 0.3, and 0.4, respectively. Our experimental observation indicates that joint training allows the generative component to implicitly provide feedback to enhance the understanding module.

 \vspace{2pt}
 \noindent\textbf{Explicit Feedback.} 
 Meanwhile, we note that the model trained in the second stage demonstrates strong generative capabilities. To leverage this, we design an efficient data filtering algorithm to select high-quality emotional images. The filtered data are subsequently used to train the model, establishing an explicit feedback loop that complements the implicit feedback from the generation component.
As shown in Alg.~\ref{alg:emo_gen_filter}, to ensure emotional relevance and semantic clarity in the generated images, we employ Emo-A and Sem-C as evaluation criteria for filtering. 
Specifically, Emo-A verifies that each image conveys the intended emotional state, while Sem-C preserves clear, recognizable content, preventing the images from becoming overly abstract or indistinct.  Detailed definitions of these metrics are provided in the experimental section. By selecting images within the top $\eta\%$ to $\delta\%$ range for both metrics, we achieve a balance between emotional expressiveness and semantic clarity. This dual constraint ensures that only the highest quality images, which effectively evoke emotions while remaining semantically interpretable, are retained for further training. 
\begin{algorithm}
        \caption{Dual Metric High-Quality Image Filtering}
            \begin{algorithmic}[1]
                \Require {generated image set $\boldsymbol{\mathcal{I}}$, CLIP model $\boldsymbol{M_{\text{CLIP}}}$, 
                \Statex \quad \quad {  } emotion classifier $\boldsymbol{M_{\text{emo}}}$, thresholds $\boldsymbol{\eta}$, $\boldsymbol{\delta}$,
                \Statex \quad \quad {  } semantic classifier} $\boldsymbol{M_{\text{sem}}}$
                \Ensure filtered image set $\boldsymbol{\mathcal{I}_{\text{filtered}}}$
                \State \textcolor{darkgray}{\text{/* compute emotion accuracy */}}
                \State $\boldsymbol{\mathbf{F}} \in \mathbb{R}^{m \times c} \leftarrow \boldsymbol{M_{\text{CLIP}}}(\boldsymbol{\mathcal{I}})$ 
                \State $\boldsymbol{\mathbf{P}_{\text{emo}}} \in \mathbb{R}^{m \times 1} \leftarrow \boldsymbol{M_{\text{emo}}}(\boldsymbol{\mathbf{F}})$
                \State \textcolor{darkgray}{\text{/* compute semantic clarity scores*/}}
                \State $\boldsymbol{\mathbf{S}_{\text{sem}}} \in \mathbb{R}^{m \times 1} \leftarrow \boldsymbol{M_{\text{sem}}}(\boldsymbol{\mathbf{F}})$
                \State \textcolor{darkgray}{\text{/* sort by emotion accuracy (Emo-A) */}}
                \State $\boldsymbol{\mathbf{R}_{\text{emo}}} \in \mathbb{R}^{m} \leftarrow \boldsymbol{\text{argsort}}(\boldsymbol{\mathbf{P}_{\text{emo}}}, \text{dim}=-1)$
                \State \textcolor{darkgray}{\text{/* sort by semantic clarity (Sem-C) */}}
                \State $\boldsymbol{\mathbf{R}_{\text{sem}}} \in \mathbb{R}^{m} \leftarrow \boldsymbol{\text{argsort}}(\boldsymbol{\mathbf{S}_{\text{sem}}}, \text{dim}=-1)$
                \State \textcolor{darkgray}{\text{/* select images within top $\boldsymbol{\eta}\%$ to $\boldsymbol{\delta}\%$  */}}
                \State $\boldsymbol{\mathcal{I}_{\text{filtered}}} \leftarrow \boldsymbol{\mathcal{I}} \left( \boldsymbol{\mathbf{R}_{\text{emo}}}[\boldsymbol{\eta}\% : \boldsymbol{\delta}\%] \cap \boldsymbol{\mathbf{R}_{\text{sem}}}[\boldsymbol{\eta}\% : \boldsymbol{\delta}\%] \right)$
                \State \textbf{Return}: $\boldsymbol{\mathcal{I}_{\text{filtered}}}$
            \end{algorithmic}
        \label{alg:emo_gen_filter}
        \end{algorithm}
        \vspace{-15pt}
        
\section{Experiments}
\label{Experiments}
\subsection{Experiments Settings}
 \noindent\textbf{Datasets.} We primarily conduct experiments on two large-scale datasets: 1) \textbf{EmoSet}~\cite{yang2023emoset} comprises 3.3 million images, with 118,102 meticulously labeled by human annotators. It comprises 8 emotion categories along with diverse attributes, including brightness, colorfulness, scene type, object class, facial expression, and human action, supporting fine-grained analysis. 2) \textbf{The Flickr and Instagram (FI)}~\cite{you2016building} dataset is a widely used image emotion dataset. It consists of 22,683 images collected from the Flickr and Instagram platforms, each labeled with one of eight emotional categories.
 
 For additional evaluation, we also test on three smaller datasets:
3) \textbf{EmotionROI}~\cite{peng2016emotions} contains 1980 emotional images from Flickr. It is annotated with six balanced categories: joy, surprise, anger, disgust, fear, and sadness. 4) \textbf{Twitter I}~\cite{you2015robust} and 5) \textbf{Twitter II}~\cite{borth2013large} are
two small datasets collected from Twitter. They have 1269 and 603
images respectively. They only contain two affective states: positive and
negative.

\vspace{5pt}
\noindent\textbf{Evaluation Metrics.} 
 Following previous works~\cite{yang2024emogen, xie2024emovit, heusel2017gans}, we employ a diverse set of metrics to evaluate our model’s understanding and generation capabilities. 1) \textbf{Accuracy}: Top-1 classification accuracy measures the model's understanding ability. 2) \textbf{FID}: Assesses distributional differences between generated and real images. 3) \textbf{Emo-A}: Evaluates alignment between generated images and intended emotions. 4) \textbf{Sem-C}: Assesses the clarity of generated image content. 5) \textbf{LPIPS}: Measures the diversity of generated emotion-driven images. 6) \textbf{Sem-D}: Estimates the richness of content linked to each emotion.

\vspace{2pt}
\noindent\textbf{Comparison Methods and Evaluation Protocols.} 
1) \textbf{Understanding}:
Following previous standard works~\cite{xu2024learning, xie2024emovit}, we evaluate our model on the large-scale EmoSet and FI datasets. We also provide extended comparisons on additional three smaller datasets: EmotionROI, Twitter I, and Twitter II. 2) \textbf{Generation}: Emotional image content generation is a newly introduced task in EmoGen~\cite{yang2024emogen}, with its only comparison benchmark EmoSet. 
All comparative generation methods are evaluated
under the same training dataset (EmoSet) and diffusion architecture
for a fair comparison. 

\vspace{2pt}
\noindent\textbf{Implementation Details.}
Our experiments are implemented in PyTorch and conducted on four NVIDIA L40S GPUs, each with 48GB of memory. In the first stage, we set the batch size to 128 and optimized the understanding component using the AdamW optimizer, with an initial learning rate of 0.0001. The AdamW parameters are configured with $\beta_1 = 0.9$, $\beta_2 = 0.999$, and a weight decay of 0.001. In the second stage, we jointly train the entire model with a learning rate of 0.01. During the fusion process, a two-layer MLP with ReLU activation is used to map the various visual representations into the input space of the diffusion model. The pre-trained emotion classifier in the emotional correlation coefficient is composed of two fully connected layers, which are trained on top of frozen CLIP features; \textcolor{b}{the coefficient is computed on the training split of the dataset.} Based on experimental results, we set the explicit feedback thresholds $\eta\%$ and $\delta\%$ to 20\% and 80\%, respectively. 
\vspace{-25pt}
\subsection{Comparison with State-of-the-Art Methods}

 \begin{table}[t]
     \footnotesize
    \caption{Comparison results (\%) for \textbf{visual emotion understanding task} on the large-scale EmoSet and FI dataset.}
     \vspace{-6pt}
        \label{tab:exp_sota}
        
        \centering
        \setlength{\tabcolsep}{10pt} 
        \footnotesize
       
        \begin{tabular}{l c c c}
            \toprule
            \textbf{Method} & \textbf{Backbone} &\textbf{EmoSet} &\textbf{FI}  \\
            \hline
            \rowcolor{gray!20} 
            \multicolumn{4}{c}{\textit{Visual Instruction Tuning}} \\
            \hline
            BLIP2~\cite{li2023blip}      &ViT-L/14  &46.79 & -        \\
            InstructBLIP \cite{dai2023instructblip} &ViT-L/14         &42.20 &-\\
            Flamingo \cite{alayrac2022flamingo} &ViT-L/14        &29.59 &-     \\
            LLaVA \cite{liu2024visual} & ViT-L/14         &44.03 &-      \\
            EmoVIT \cite{xie2024emovit} &ViT-L/14        &83.36 &-     \\
            \hline
            \rowcolor{gray!20} 
            \multicolumn{4}{c}{\textit{Supervised Emotion Recognition}}\\
            \hline
            MDAN \cite{xu2022mdan} &ResNet-101 &75.75 & 76.41\\
            \hline
            CoOp \cite{zhou2022learning} &ViT-B/32 & 76.19 & 78.66  \\
            CoCoOp \cite{zhou2022conditional} &ViT-B/32 & 80.31 &77.46 \\
            SimEmotion \cite{deng2022simemotion} &ViT-B/32 & 79.06 &80.33 \\
            PT-DPC \cite{deng2024learning} &ViT-B/32  &77.13 &78.07\\
            MASANet \cite{cen2024masanet} &ViT-B/32 & -&79.16 \\
            MVP \cite{xu2024learning}&ViT-B/32 & 81.92 &82.76\\
            \hline
            \rowcolor{blue!10}
            \textbf{UniEmo} &ViT-B/32  & \textbf{83.52} &\textbf{85.22}\\
            \rowcolor{blue!10}
            \textbf{UniEmo} &ViT-B/16  & \textbf{84.54} &\textbf{86.34}\\
             \rowcolor{blue!10}
            \textbf{UniEmo} &ViT-L/14  & \textbf{85.30} &\textbf{87.65}\\
            \bottomrule
        \end{tabular}
        \vspace{-8pt}
        \end{table}
        
         \begin{table}[t] 
    \caption{Additionally main results for \textbf{visual emotion understanding task} on other small datasets.}
        \label{tab:main_results}
        \vspace{-6pt}
        \centering
        \renewcommand\arraystretch{1.18}
        \setlength\tabcolsep{3pt}
        \footnotesize
        \begin{tabular}{lccccc}
            \toprule
            \textbf{Method} & \textbf{Backbone } &\textbf{EmotionROI} & \textbf{Twitter I} & \textbf{Twitter II}   \\
             \hline
            CoOp \cite{zhou2022learning}& ViT-B/32 &68.48 &89.05 &83.78\\
            CoCoOp \cite{zhou2022conditional}& ViT-B/32    &71.09 &91.63 &82.94 \\
            PT-DPC \cite{deng2024learning}&  ViT-B/32   &69.70 &90.94 &82.50 \\
            SimEmotion \cite{deng2022simemotion}&  ViT-B/32 &70.54 &89.76 &84.21 \\
            MVP \cite{xu2024learning}&  ViT-B/32   &71.89 &92.03 &88.21 \\
            \hline
            \rowcolor{blue!10}
            \textbf{UniEmo} & ViT-B/32 & \textbf{74.25} &    \textbf{93.98} &   \textbf{89.94} \\
             \rowcolor{blue!10}
            \textbf{UniEmo} & ViT-B/16 & \textbf{75.57} &    \textbf{94.75}  &    \textbf{90.69} \\
            \rowcolor{blue!10}
            \textbf{UniEmo} & ViT-L/14 & \textbf{76.78} &    \textbf{95.63} &    \textbf{91.68} \\
            \bottomrule
        \end{tabular}
        \vspace{-2em}
    \end{table}
    \begin{table}[t]
    \caption{\textbf{Zero-shot performance} comparison for \textbf{visual emotion understanding task} on the FI dataset.  Following the experimental protocol of EmoVIT~\cite{xie2024emovit}, all methods are pretrained on the EmoSet dataset,  highlighting the generalization capabilities of our UniEmo.}
     \vspace{-6pt}
        \label{tab:exp_sota_FI_zero}
        \centering
        \setlength{\tabcolsep}{15pt} 
        \footnotesize
        
        \begin{tabular}{l c c}
            \toprule
            \textbf{Method} & \textbf{Backbone} & \textbf{Accuracy (\%)}  \\
            \hline
            BLIP2~\cite{li2023blip} &ViT-L/14 & 57.72 \\
            InstructBLIP \cite{dai2023instructblip}&ViT-L/14 & 37.97 \\
            Flamingo \cite{alayrac2022flamingo}&ViT-L/14 &14.91 \\
            LLaVA \cite{liu2024visual}&ViT-L/14 & 56.04 \\
            EmoVIT \cite{xie2024emovit}&ViT-L/14 & 68.09 \\
            \hline
            CoOp   \cite{zhou2022learning} &ViT-B/32          &60.52   \\
            CoCoOp \cite{zhou2022conditional} &ViT-B/32       &    61.98\\
            MVP \cite{xu2024learning}  & ViT-B/32        &65.73  \\
            \hline
            \rowcolor{blue!10}
            \textbf{UniEmo} & ViT-B/32 & \textbf{68.32} \\
             \rowcolor{blue!10}
            \textbf{UniEmo} & ViT-B/16 & \textbf{70.05} \\
            \rowcolor{blue!10}
            \textbf{UniEmo} & ViT-L/14 & \textbf{71.22} \\
            \bottomrule
        \end{tabular}
        \vspace{-15pt}
    \end{table} 
    
      \begin{table*}
        \caption{\textbf{Performance comparison on visual emotion generation task} on the EmoSet dataset. We follow the settings established in EmoGen~\cite{yang2024emogen}, where all compared methods are trained on EmoSet and adopt the same architecture (stable-diffusion-v1-5 for SD 1.5 and stable-diffusion-xl-base-1.0 for SD XL). $^\dagger$~denotes results reproduced from the official implementation in EmoGen~\cite{yang2024emogen}.}
  	\label{tab:exp_sota_gen}
        \vspace{-6pt}
        \centering
        \renewcommand\arraystretch{1.18}
        \setlength\tabcolsep{14pt}
        \footnotesize
        \begin{tabular}{lccccccccc}
            \toprule
            \textbf{Method} &\textbf{Diffusion Architecture} &\textbf{FID $\downarrow$} & \textbf{LPIPS $\uparrow$} & \textbf{Emo-A $\uparrow$} & \textbf{Sem-C $\uparrow$} & \textbf{Sem-D $\uparrow$}  \\
            \midrule
            Stable Diffusion~\cite{rombach2022high}&SD 1.5 & 44.05 & 0.687 & 70.77\% & 0.608 & 0.0199  \\
            DreamBooth~\cite{ruiz2023dreambooth}&SD 1.5  & 46.89 & 0.661 & 70.50\% & 0.614 &  0.0178 \\
            Textual Inversion~\cite{gal2022image}&SD 1.5 & 50.51 & 0.702 & 74.87\% & 0.605 & 0.0282  \\
            EmoGen~\cite{yang2024emogen}&SD 1.5 & 41.60 & 0.717 & 76.25\% &0.633 &  0.0335 \\
            \rowcolor{blue!10}
            \textbf{UniEmo}&SD 1.5  & \textbf{27.73} & \textbf{0.793}\ & \textbf{79.66\%} & \textbf{0.640} & \textbf{0.0383} \\
            \hline
            \rowcolor{gray!20} 
            \multicolumn{7}{c}{\textit{Generalization across Architectures}} \\
            \hline
            Stable Diffusion XL$^\dagger$~\cite{podell2023sdxl}&SD XL&41.57  &0.721  &76.18\% &0.628 &0.0298   \\
            EmoGen$^\dagger$~\cite{yang2024emogen} &SD XL &37.23  &0.753  &78.92\% &0.638  &0.0357  \\
            \rowcolor{blue!10}
            \textbf{UniEmo}&SD XL  & \textbf{26.61} & \textbf{0.807}\ & \textbf{81.74\%} & \textbf{0.642} & \textbf{0.0392} \\
            \bottomrule
        \end{tabular}
        \vspace{-8pt}
     \end{table*}
     
      \begin{figure*}
        \centering
        \includegraphics[width=1.0\linewidth]{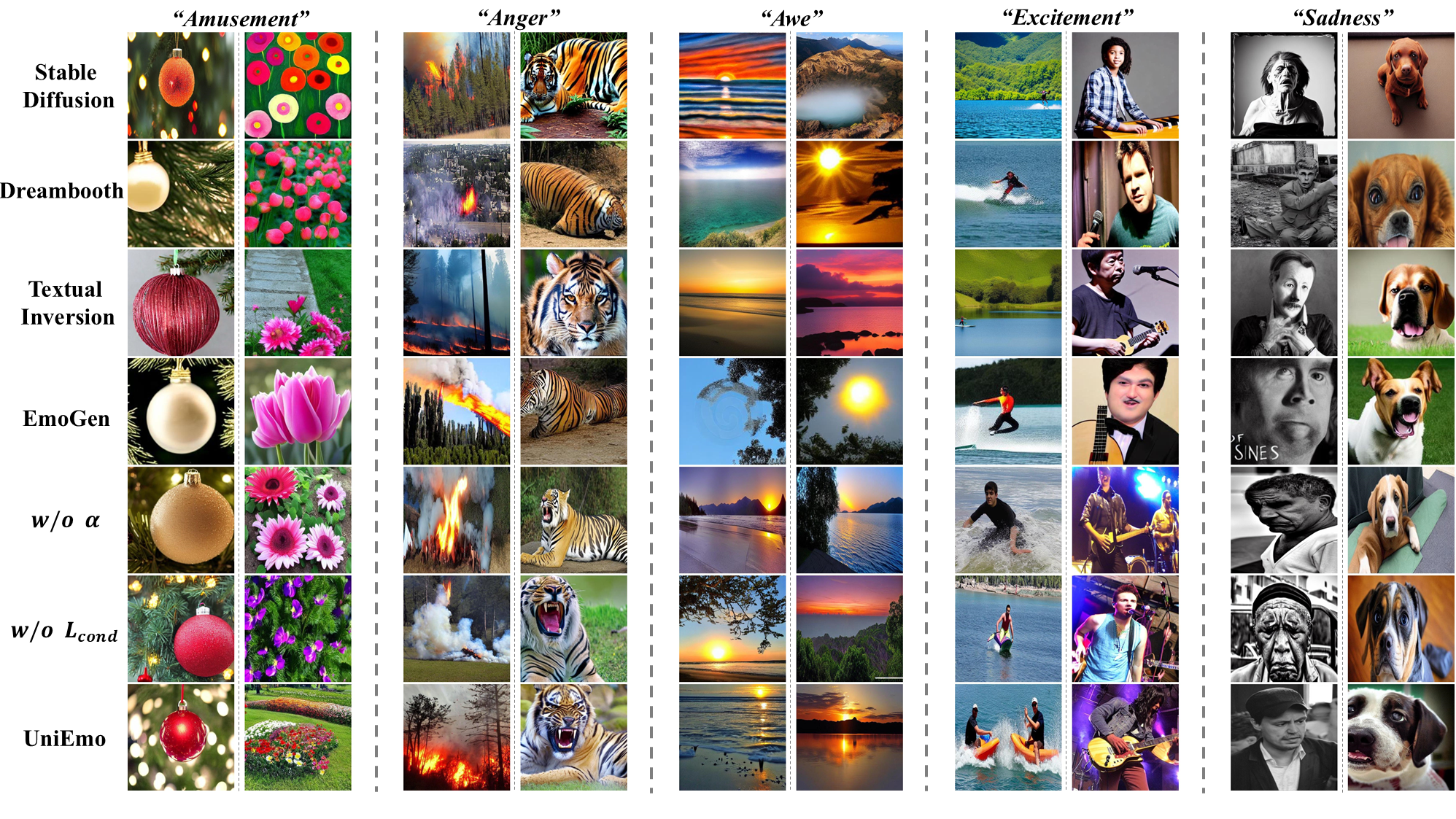}
        \caption{Qualitative comparison with the state-of-the-art emotional generation methods. To facilitate a more comprehensive comparison, we examine generated images with consistent semantic themes. The settings w/o $\alpha$ and $L_{cond}$ represent the results where the emotional correlation coefficient and emotional condition loss are removed, respectively.}
        \label{fig:vis}
        \vspace{-20pt}
    \end{figure*}
    
\noindent\textbf{Visual Emotion Understanding.}
As shown in Tab.~\ref{tab:exp_sota}, UniEmo demonstrates consistently superior supervised performance on the large-scale EmoSet and FI datasets. Under the ViT-B/32 backbone, UniEmo attains 83.52\% on EmoSet and 85.22\% on FI, outperforming strong baselines such as CoCoOp, which achieves 80.31\% and 77.46\%, and MVP, which reaches 81.92\% and 82.76\%, respectively. Notably, the performance gain on FI exceeds seven percentage points over CoCoOp, evidencing the robustness of our approach in cross-domain evaluation. When scaled to larger backbones, UniEmo maintains its advantage: with ViT-B/16, it achieves 84.54\% on EmoSet and 86.34\% on FI; with ViT-L/14, the performance further improves to 85.30\% and 87.65\%. This not only surpasses the previous best reported by EmoVIT, which records 83.36\% on EmoSet, but also establishes a new state-of-the-art across both benchmarks. 
Evaluations on smaller benchmarks, reported in Tab.~\ref{tab:main_results}, reveal consistent advantages of UniEmo under the ViT-B/32 backbone. On EmotionROI, UniEmo achieves 74.25\%, outperforming CoCoOp at 71.09\% and SimEmotion at 70.54\%, with an absolute gain of over three points. On Twitter I, UniEmo reaches 93.98\%, exceeding CoCoOp (91.63\%) and SimEmotion (89.76\%), while on Twitter II, it attains 89.94\%, surpassing CoCoOp (82.94\%) and SimEmotion (84.21\%) by notable margins. These results demonstrate that UniEmo delivers consistent improvements of approximately 3--7 percentage points across small-scale benchmarks when compared under the same backbone. Moreover, with larger backbones such as ViT-B/16 and ViT-L/14, UniEmo further pushes performance to 94.75\% and 95.63\% on Twitter I and up to 91.68\% on Twitter II, establishing new state-of-the-art results. Together, these findings highlight UniEmo’s robustness in limited-data scenarios and confirm that its gains are not confined to large-scale benchmarks but extend effectively to smaller and more domain-specific datasets.

To further assess cross-dataset generalization, we follow the EmoVIT~\cite{xie2024emovit} protocol and conduct zero-shot evaluations on FI, with results summarized in Tab.~\ref{tab:exp_sota_FI_zero}. In this setting, all models are pretrained solely on EmoSet without fine-tuning on FI. UniEmo achieves 68.32\% with the ViT-B/32 backbone, outperforming MVP by more than two points and slightly surpassing EmoVIT. With the ViT-L/14 backbone, UniEmo reaches 71.22\%, setting a new state-of-the-art in zero-shot emotion understanding. These results further demonstrate its robust generalization capability, effectively avoiding dataset-specific biases.\textcolor{b}{The consistently stronger emotion understanding performance, both in-domain and under cross-domain shift, can be largely attributed to two design choices. First, UniEmo leverages learnable expert queries to disentangle scene-level context from object-centric cues, which mitigates spurious background correlations and promotes attention to semantically grounded emotion evidence. Second, the unified understanding–generation formulation encourages representations that are not only discriminative for recognition but also informative for conditional synthesis, effectively acting as a regularizer and improving robustness across domains.}

\noindent\textbf{Visual Emotion Generation.}
    To evaluate the visual emotion generation capabilities of UniEmo, we perform a fair comparison using the metrics established in EmoGen~\cite{yang2024emogen}. The results in Tab.~\ref{tab:exp_sota_gen} demonstrate that UniEmo significantly outperforms existing state-of-the-art methods in the visual emotion generation task across multiple evaluation metrics. UniEmo achieves superior performance in terms of FID, LPIPS, and Emo-A, indicating its ability to generate high-quality, perceptually diverse, and emotion-consistent images. Moreover, the model also excels in Sem-C and Sem-D, showcasing its capacity to produce semantically coherent and rich visual outputs. \textcolor{b}{The improvements are largely attributable to a more informative conditioning signal derived from our understanding branch, which grounds diffusion sampling in semantically meaningful emotion cues and mitigates semantic drift. Specifically, the scene and object expert queries capture complementary affective evidence, including global context and salient entities, while the correlation-aware fusion adaptively emphasizes attributes that are most relevant to the target emotion. Moreover, the condition loss promotes better separation among easily confused emotions, thereby improving emotion controllability without compromising visual diversity.}
    \vspace{-10pt}
    
  \begin{figure*}
        \centering
        \begin{minipage}[t]{0.32\linewidth}
            \centering
            \includegraphics[width=\linewidth]{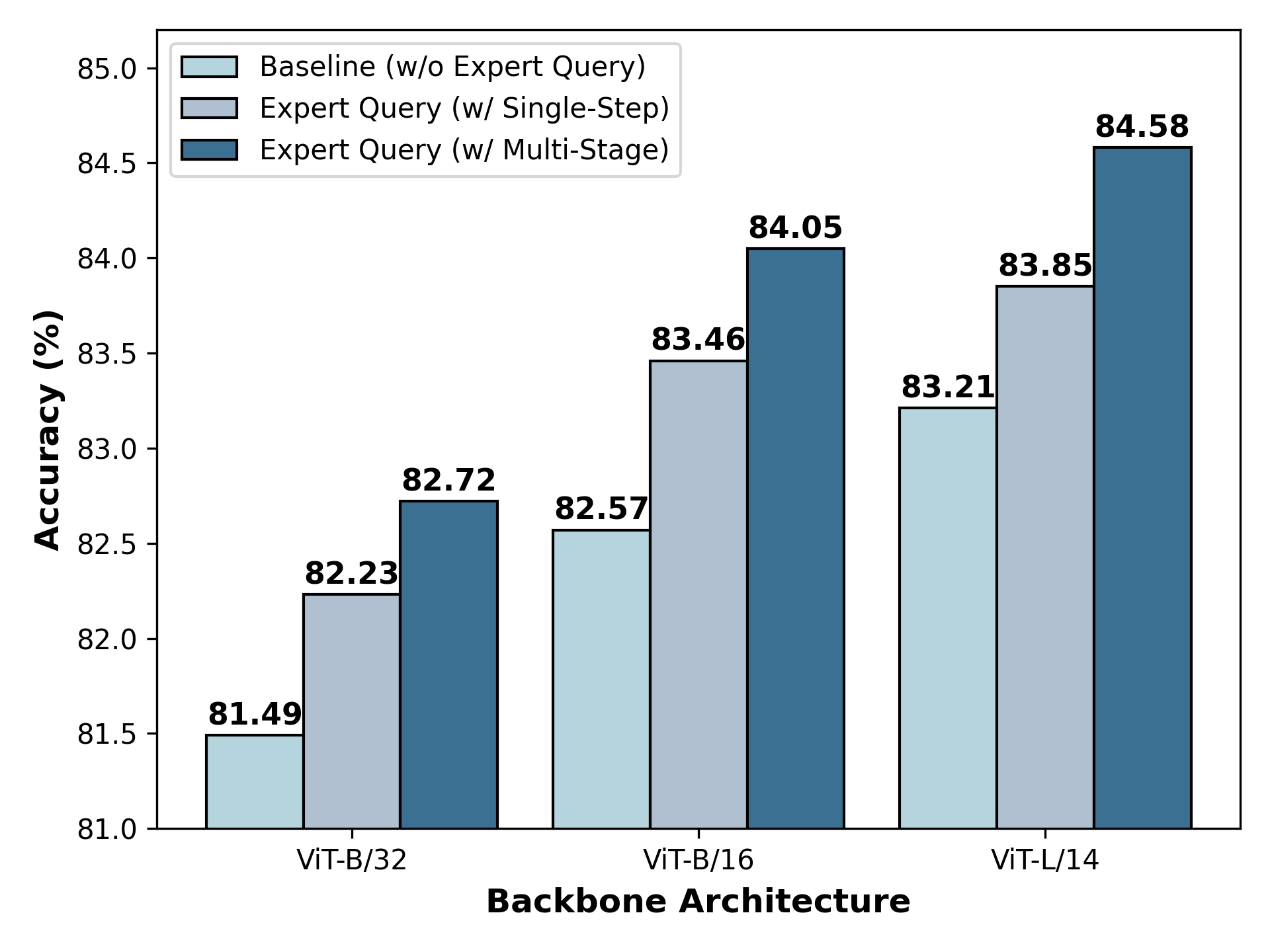}
            \vspace{-16pt}
            \caption{\textbf{Ablation study on expert query in visual emotion understanding}  task across various backbone architectures.}
            \label{fig:Expert_queries}
        \end{minipage}
        \hfill
        \begin{minipage}[t]{0.32\linewidth}
            \centering
            \includegraphics[width=\linewidth]{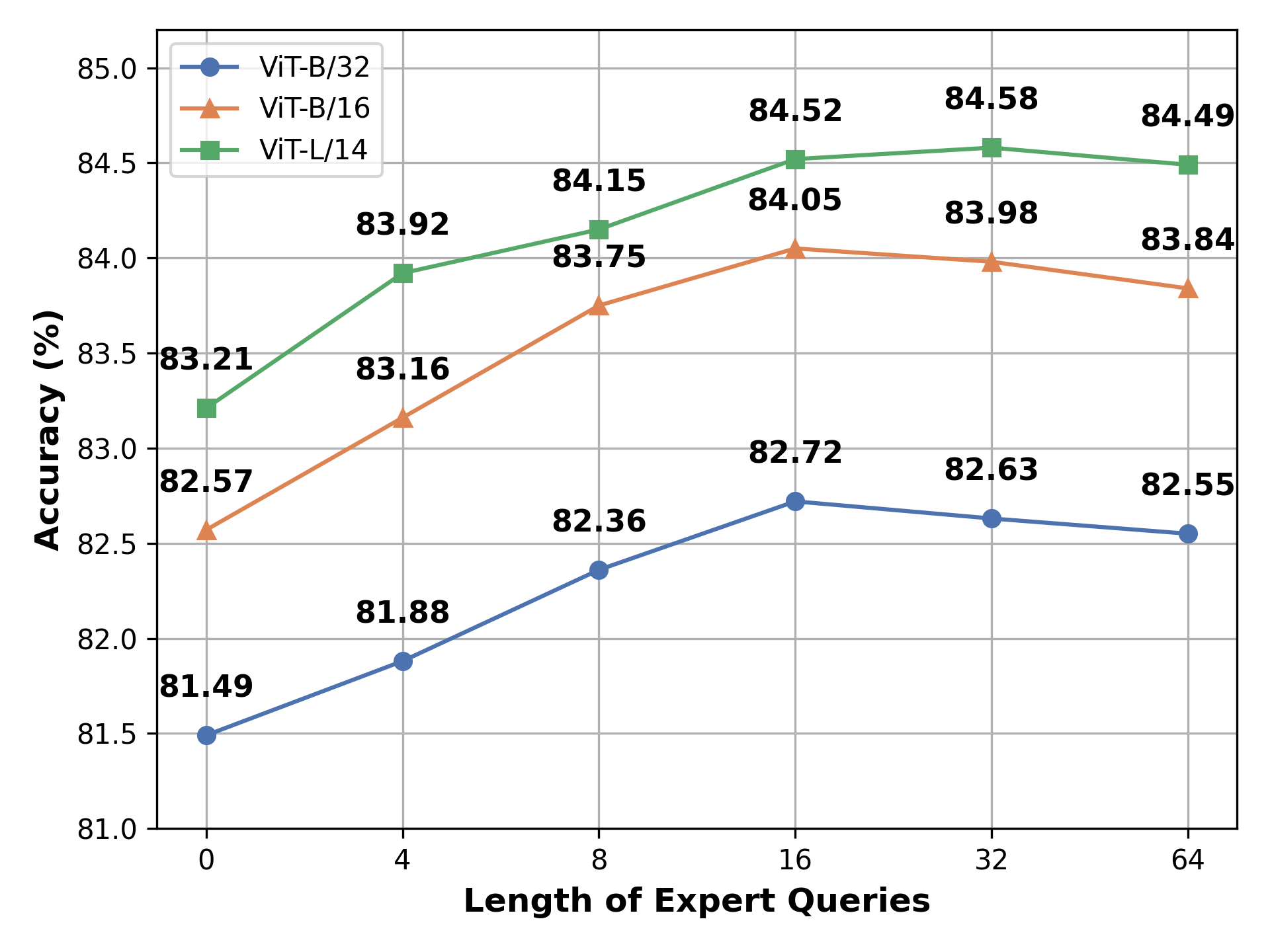}
            \vspace{-16pt}
            \caption{\textbf{Ablation study on expert query length} in visual emotion understanding task across various  architectures.}
            \label{fig:Length_queries}
        \end{minipage}
        \hfill
        \begin{minipage}[t]{0.32\linewidth}
            \centering
            \includegraphics[width=\linewidth]{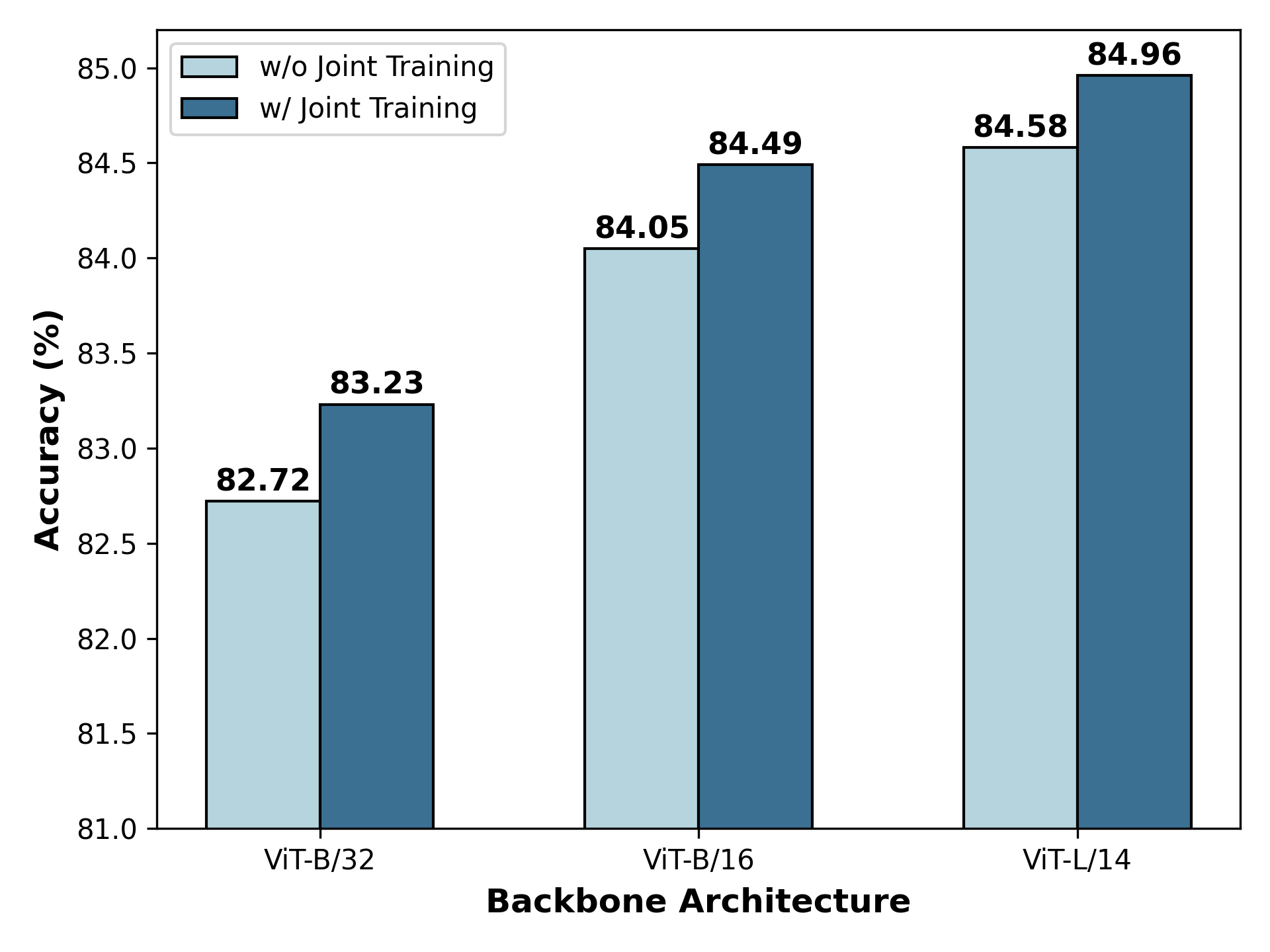}
            \vspace{-16pt}
            \caption{\textbf{Ablation study on implicit feedback} in visual emotion understanding task across various  architectures.}
            \label{fig:joint_training}
        \end{minipage}
        \vspace{-5pt}
    \end{figure*}
    
  \begin{table*}
                \caption{\textbf{Ablation study on the expert query, emotional correlation coefficient, and emotional condition loss} in visual emotion generation. The table evaluates the use of class (Cls), patch (Patch) tokens, and expert queries (Scene and Object). Superscript * denotes the use of the emotional correlation coefficient. $L_{ce}$ is cross-entropy loss; $L_{cond}$ is the proposed emotional condition loss in this paper.}
                 
        \label{tab:ablition_gen}
        \vspace{-6pt}
        \centering
        \renewcommand\arraystretch{1.18}
        \footnotesize
        \setlength{\tabcolsep}{20pt}
        \begin{tabular}{lcccccccc}
            \toprule
            \textbf{Method} & \textbf{FID $\downarrow$} & \textbf{LPIPS $\uparrow$} & \textbf{Emo-A $\uparrow$} & \textbf{Sem-C $\uparrow$} & \textbf{Sem-D $\uparrow$}  \\
		  \hline
            \rowcolor{gray!20} 
            \multicolumn{6}{c}{\textit{Combination of Expert Query}} \\
            \hline
		  Cls & 30.02 &0.763 & 71.35\% & 0.621& 0.0331\\
		  Cls + Object & 29.99&	0.769	&73.12\%&	0.625	&0.0337 \\
  		Cls + Scene & 30.30&	0.771&	73.10\%&	\textbf{0.627}	&0.0342  \\
		  Cls + Patch & 29.86	&0.764&	70.62\%&	0.617&	0.0335 \\
		  Cls + Object + Scene  & \textbf{29.77}	&\textbf{0.775}&\textbf{74.07\%}&	   0.625&	\textbf{0.0351}\\
            Cls + Patch + Object + Scene  & 29.83&	0.770&	72.77\%&0.621&	0.0339 \\
            \hline
            \rowcolor{gray!20} 
            \multicolumn{6}{c}{\textit{Emotional Correlation Coefficient}} \\
            \hline
            Cls + Object*  & 29.21&	0.778&	75.62\%&	0.630&	0.0352 \\
            Cls + Scene*  & 29.13&	0.777&	75.44\%&	0.631&	0.0349 \\
            Cls + Object* + Scene*  & \textbf{28.56}&\textbf{0.784}	&\textbf{77.89\%}&	\textbf{0.633}	&\textbf{0.0358} \\
            \hline
            \rowcolor{gray!20} 
            \multicolumn{6}{c}{\textit{Emotional Condition Loss}} \\
            \hline
            Cls + Object* + Scene* + $L_{\text{ce}}$ &28.08	&0.787&	78.97\%&	0.635&	0.0357 \\
            Cls + Object* + Scene* + $L_{\text{cond}}$ & \textbf{27.87}&	\textbf{0.790}&	\textbf{79.77\%}&	\textbf{0.637}&	\textbf{0.0367} \\
            \bottomrule
        \end{tabular}
        \vspace{-10pt}
    \end{table*}

     
     In Fig.~\ref{fig:vis}, we qualitatively compare UniEmo with existing generation methods across five emotions. To facilitate intuitive comparisons, we evaluated generated images with consistent semantic themes \textcolor{b}{(e.g., tigers, puppies, and lakes).} The final row showcases the outputs from UniEmo, which demonstrate a higher degree of realism and emotional alignment when compared to the preceding methods. Specifically, the generated images by UniEmo are more contextually coherent, with vivid details that effectively convey the target emotions. For example, the tigers in the \textit{anger} category display intense expressions and dynamic postures, while other methods generate images with ambiguous emotional cues. \textcolor{b}{Such fine-grained affective details are better captured by our object-query conditioning, which emphasizes salient entities and their emotion-relevant expressive patterns.} In the \textit{amusement} category, our method generates vibrant, diverse flowers with rich colors and textures, closely matching the intended emotional theme. In contrast, Dreambooth~\cite{ruiz2023dreambooth} and EmoGen~\cite{yang2024emogen} generate more repetitive and less varied flowers. \textcolor{b}{This increased diversity is attributed to the correlation-weighted fusion that down-weights emotionally neutral attributes, together with the complementary scene--object conditioning, which helps reduce repetitive patterns and improves sample diversity.} These results show that UniEmo generates semantically rich and visually appealing images.
     
     \begin{table*} 
    \caption{Ablation study on the effectiveness of the data filtering algorithm as explicit feedback. The baseline employs no data filtering algorithm, in which case it lacks explicit feedback, while other configurations apply it using different criteria.}
        \label{tab:ablition_explicit}
        \vspace{-6pt}
        \centering
        \renewcommand\arraystretch{1.18}
        \setlength\tabcolsep{12pt}
        \footnotesize
        \begin{tabular}{lccccccccc}
            \toprule
            \textbf{Method} & \textbf{Accuracy $\uparrow$} &\textbf{FID $\downarrow$} & \textbf{LPIPS $\uparrow$} & \textbf{Emo-A $\uparrow$} & \textbf{Sem-C $\uparrow$} & \textbf{Sem-D $\uparrow$}  \\
            \hline
            Baseline (w/o Explicit Feedback) & 84.96\%&	27.87&	0.790&	\textbf{79.77\%}&	0.637&	0.0367\\
            Explicit Feedback (w/ Emo-A) & 85.10\%	&27.75&	0.781&	75.80\%	&0.611&	0.0361\\
            Explicit Feedback (w/ Sem-C) & 84.99\%	&27.93	&0.783&	75.10\%&	0.615&	0.0380  \\
            Explicit Feedback (w/ Emo-A\&Sem-C)&\textbf{ 85.30\%}	&\textbf{27.73}	&\textbf{0.793}	&79.66\%	&\textbf{0.640}&	\textbf{0.0383} \\
            \bottomrule
        \end{tabular}
        \vspace{-1em}
    \end{table*}

    \begin{table*}[t] 
    \caption{Ablation study on the impact of different threshold settings $\eta$ and $\delta$ in the data filtering algorithm.}
        \label{tab:ablition_threshold}
        \vspace{-6pt}
        \centering
        \renewcommand\arraystretch{1.18}
        \setlength\tabcolsep{14pt}
        \footnotesize
        \begin{tabular}{lccccccccc}
            \toprule
            \textbf{Method} & \textbf{Accuracy $\uparrow$} &\textbf{FID $\downarrow$} & \textbf{LPIPS $\uparrow$} & \textbf{Emo-A $\uparrow$} & \textbf{Sem-C $\uparrow$} & \textbf{Sem-D $\uparrow$}  \\
            \hline
            Baseline (w/o Explicit Feedback) & 84.96\%&	27.87&	0.790&	\textbf{79.77\%}&	0.637&	0.0367\\
            Explicit Feedback (0 - 100\%) & 85.17\%	&27.92&	0.778&	75.45\%	&0.623&	0.0365\\
            Explicit Feedback (10\% - 90\%) & 85.13\%	&27.76	&0.782&	75.71\%&	0.624&	0.0371  \\
            Explicit Feedback (50\% - 80\%)&85.16\%	&27.96	&0.783	&76.54\%	&0.620&	0.0370\\
            Explicit Feedback (50\% - 100\%)&85.05\%	&27.84	&0.775	&76.98\%	&0.625&	0.0375 \\
            Explicit Feedback (20\% - 80\%)&\textbf{ 85.30\%}	&\textbf{27.73}	&\textbf{0.793}	&79.66\%	&\textbf{0.640}&	\textbf{0.0383} \\
            \bottomrule
        \end{tabular}
        \vspace{-1.5em}
    \end{table*}
    
    \vspace{-10pt}
     \subsection{Ablation Study}
     \vspace{-3pt}
     We primarily conduct ablation studies on the large-scale EmoSet dataset to validate the effectiveness of our different modules. \textcolor{b}{We first briefly summarize our overall ablation workflow to clarify the experimental procedure: As described in Sec.~\ref{sec: 3-3}, we first train the emotion understanding component in Stage~1, and then jointly train the understanding and generation components in Stage~2. Thus, we first evaluate the expert queries in isolation for emotion understanding in Stage~1, without the influence of other components (see \textit{\textbf{C1}} below). We then enable the generation branch and conduct Stage~2 joint training, which provides implicit feedback from generation to understanding but does not yet include explicit feedback. Under this setting, we ablate key factors affecting generation, including expert queries (see \textit{\textbf{C2}}) and the emotional correlation coefficient with the emotional condition loss (see \textit{\textbf{C3}}), and we further assess the resulting implicit enhancement on understanding performance (see \textit{\textbf{C4}}). Finally, building on the above setting, we incorporate explicit feedback and examine its impact on both emotion understanding and generation performance (see \textit{\textbf{C4, C5, C6}}). We conclude this section with a summary and discussion (see \textit{\textbf{C7}})}.
     
     \noindent\textit{\textbf{C1:~}}\textbf{The Effect of Expert Query in Visual Emotion Understanding.}
     Fig.\ref{fig:Length_queries} explores the impact of query length, suggesting that moderate lengths enhance representation, while overly long queries may introduce complexity. \textcolor{b}{This trend reflects an accuracy–efficiency trade-off: moderate query capacity is sufficient to encode diverse emotional semantics, whereas excessively long queries can introduce redundant interactions and dilute attention. Based on this, we fix the query length to the best-performing setting and conduct ablations on the expert components.}
     As shown in Fig.~\ref{fig:Expert_queries}, incorporating expert queries consistently improves the performance. Furthermore, we compare two different integration strategies: 1) \textbf{Single-step}, where expert queries are introduced at the input stage, and 2) \textbf{Multi-stage}, as proposed in this paper, where expert queries are progressively introduced at different layers. The results demonstrate that the multi-stage strategy outperforms the single-step, highlighting its superiority in refining feature representations. \textcolor{b}{This is because multi-stage queries can capture global emotional context in earlier layers and then progressively focus on localized, emotion-relevant evidence through deeper cross-token interactions, yielding more robust representations.}
Additional visualizations are provided in Section~\ref{Qualitative}, demonstrating that our expert queries can effectively attend to \textit{scene-level} and \textit{object-level} information. 

 \vspace{2pt}
\noindent\textit{\textbf{C2:~}}\textbf{Understanding Improves Generation via Expert Queries.}
 As presented in Tab.~\ref{tab:ablition_gen}, using the Cls token with two expert queries yields notable performance improvements across multiple metrics. Specifically, the combination of the Cls token with two expert queries shows enhanced results in LPIPS, Emo-A, Sem-C, and Sem-D scores, along with a reduced FID, indicating better quality and diversity in generated images. 
\textcolor{b}{This is because cls provides a compact global summary, while the scene and object queries capture complementary emotion-relevant cues, forming a richer yet focused conditioning signal for diffusion. In contrast, adding dense patch tokens can introduce redundant low-level textures and background variations that are not strongly emotion-discriminative. Since the scene and object queries have already distilled much of the emotion-related patch information through prior interactions, explicitly including raw patch tokens is largely redundant and can reintroduce noise, making the conditioning less focused and weakening controllability and semantic coherence.} Overall, these findings indicate that the expert queries effectively capture more nuanced and detailed features, thereby enhancing the generative capabilities.

 \begin{figure*}
            \centering
            \includegraphics[width=0.95\linewidth]{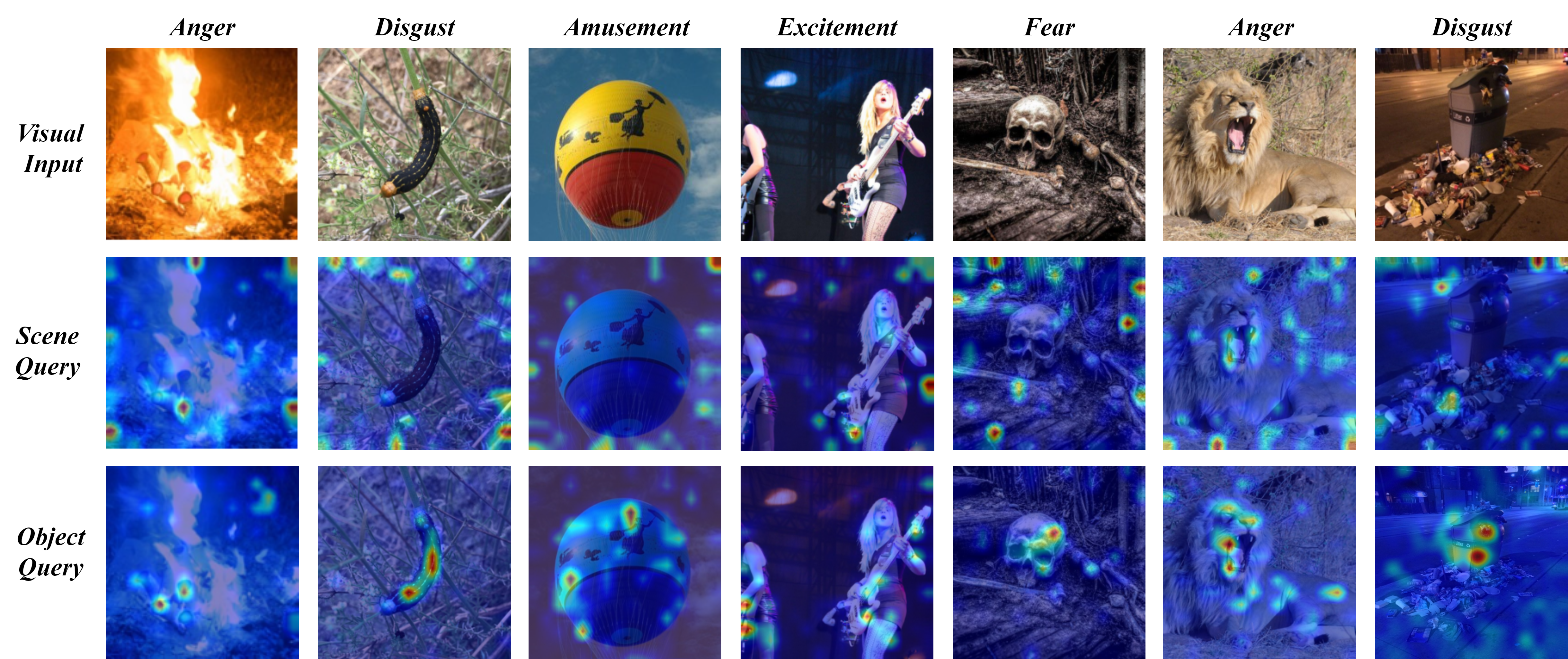}
            \vspace{-6pt}
            \caption{Visualization of the attention maps for expert queries. It demonstrates that our expert queries can effectively attend to scene and object information within emotional images, respectively.}
            \label{fig:query_merged}
            \vspace{-10pt}
\end{figure*}

 \vspace{2pt}
\noindent\textit{\textbf{C3:~}}\textbf{The Effect of Emotional Correlation Coefficient and Emotional Condition Loss.}
As shown in Tab.~\ref{tab:ablition_gen}, incorporating the emotional correlation coefficient significantly improves performance under the same token combination. This demonstrates that it improves fusion by emphasizing emotion-relevant features, resulting in more realistic image generation. \textcolor{b}{The coefficient acts as an adaptive weighting prior that down-weights emotionally neutral attributes and up-weights those strongly associated with the target emotion, which helps reduce off-emotion artifacts while preserving semantically meaningful details.}
Additionally, emotional condition loss outperforms standard cross-entropy, notably improving the Emo-A score by better aligning fused features with the target emotion space. \textcolor{b}{Compared with cross-entropy, the condition loss explicitly encourages better separation among easily confused emotions, making the conditioning signal more discriminative and thus improving emotion controllability during diffusion sampling.}
Meanwhile, the qualitative analysis in Fig.~\ref{fig:vis} further illustrates their importance: \textcolor{b}{without these coefficients, the generated images tend to exhibit weaker emotion cues and less coherent details, resulting in poorer emotion alignment.}

 \vspace{2pt}
\noindent\textit{\textbf{C4:~}}\textbf{Generation Enhances Understanding via Implicit and Explicit Feedback.}
As shown in Fig.~\ref{fig:joint_training}, joint training in the second stage enhances the model's understanding by incorporating generation tasks across different backbone architectures. This demonstrates that the generation component provides implicit feedback, optimizing discriminative emotional representations and consistently improving understanding. \textcolor{b}{Since generation must reconstruct visually coherent images that match the target emotion, it encourages the model to preserve semantically grounded emotion cues, thereby regularizing the encoder toward more informative and transferable features than purely discriminative training.} Furthermore, we conduct an ablation study to evaluate the effectiveness of the data filtering algorithm as explicit feedback for the understanding component. As shown in Tab.~\ref{tab:ablition_explicit}, incorporating this filtering mechanism markedly improves the model's performance in both emotional understanding and generation. Additionally, employing a dual-metric filtering strategy using Emo-A and Sem-C demonstrates superior results, outperforming single-metric approaches by effectively balancing semantic and emotional consistency in the generated output.

    \vspace{2pt}
\noindent\textit{\textbf{C5:~}}\textbf{Different Threshold Settings in the Data Filtering Algorithm.}
In Table~\ref{tab:ablition_threshold}, we present an ablation study to evaluate the impact of different threshold settings $\eta$ and $\delta$ on the performance of the data filtering algorithm. The baseline method, which lacks explicit feedback, achieves an accuracy of 84.96\%, with an FID of 27.87, and a high Emo-A score of 79.77\%. However, introducing explicit feedback across various threshold ranges consistently improves several metrics. Notably, the configuration with explicit feedback set between 20\% and 80\% yields the highest accuracy (85.30\%), the lowest FID (27.73), and the best LPIPS score (0.793). Additionally, this configuration achieves the highest scores in Sem-C (0.640) and Sem-D (0.0383), indicating improved semantic consistency and diversity. Although other configurations also improve performance over the baseline, the 20\%-80\% setting consistently outperforms them across most metrics. \textcolor{b}{This trend reflects a trade-off between feedback quality and diversity. }

\begin{table}[t] 
    \caption{Ablation study of the proportion of generated images in explicit feedback during training.}
        \label{tab:ablition_proportion}
        \vspace{-6pt}
        \centering
        \renewcommand\arraystretch{1.18}
        \setlength\tabcolsep{3pt}
        \footnotesize
        \begin{tabular}{lccccccccc}
            \toprule
            \textbf{Method} & \textbf{Accuracy $\uparrow$} &\textbf{FID $\downarrow$} & \textbf{LPIPS $\uparrow$} & \textbf{Emo-A $\uparrow$} & \textbf{Sem-C $\uparrow$} & \textbf{Sem-D $\uparrow$}  \\
            \hline
            5\% & 85.01\%&	27.85&	0.787&	77.56\%&	0.634&	0.0358\\
            10\%  & 85.12\%	&27.81&	0.791&	77.91\%	&0.635&	0.0369\\
            20\% & 85.30\%	&\textbf{27.73}	&\textbf{0.793}&	\textbf{79.66\%}&	\textbf{0.640}&	0.0383  \\
            30\% &\textbf{85.34\%}	&27.75	&0.792	&78.34\%	&0.640&	\textbf{0.0389}\\
            \bottomrule
        \end{tabular}
        \vspace{-2em}
\end{table}

 \vspace{2pt}
\noindent\textit{\textbf{C6:~}}\textbf{The Proportion of Generated Images During Training.}
In Tab.~\ref{tab:ablition_proportion}, we investigated the impact of the proportion of generated images on model performance.  Our analysis revealed that varying the percentage of generated images significantly influences key performance metrics. Specifically, we observed that increasing the proportion of generated images generally enhances model accuracy and perceptual quality, as indicated by improvements in metrics such as Accuracy, FID, and LPIPS. However, this improvement comes at the cost of increased computational resources and processing time. \textcolor{b}{Overall, this reflects a trade-off: a moderate amount of generated data improves robustness by expanding emotion-relevant variations, whereas too much can introduce noisy samples and dilute supervision from real data.} Under this balance, the 20\% setting provides a favorable compromise.

 \vspace{2pt}
\noindent\textit{\textbf{C7:~}}\textbf{Conclusion and Discussion.}
\textcolor{b}{We summarize the roles of the interacting components in UniEmo and distinguish core design choices from auxiliary enhancements:}

\noindent\textcolor{b}{\textbf{1) Expert Queries (Core Design)} provide a structured decomposition by capturing complementary scene-level context and object-level cues, yielding semantically grounded representations that improve both emotional understanding and generation (Fig.~\ref{fig:Expert_queries}, Fig.~\ref{fig:Length_queries}, Tab.~\ref{tab:ablition_gen}).}
    
\noindent\textcolor{b}{\textbf{2) Emotional Correlation Coefficient and Emotional Condition Loss (Core Designs).} These two designs are crucial when fusing the understanding-derived queries into the generation conditioning, substantially strengthening the effectiveness of the conditioning signal and thereby markedly improving UniEmo’s emotional generation performance (Tab.~\ref{tab:ablition_gen}, Fig.~\ref{fig:vis}).}

\noindent\textcolor{b}{\textbf{3) Implicit Feedback (Auxiliary Enhancement).} During Stage~2 joint training, the generation objective encourages the model to retain semantically grounded emotion cues, which regularizes the encoder and further improves emotion understanding beyond purely discriminative training (Fig.~\ref{fig:joint_training}).}

\noindent\textcolor{b}{\textbf{4) Explicit Feedback (Core Design).} Incorporating the proposed filtering mechanism as explicit feedback consistently improves both emotion understanding and generation performance (Tab.~\ref{tab:ablition_explicit}, Tab.~\ref{tab:ablition_threshold}, Tab.~\ref{tab:ablition_proportion}).}

  \begin{figure*}
            \centering
            \includegraphics[width=0.95\linewidth]{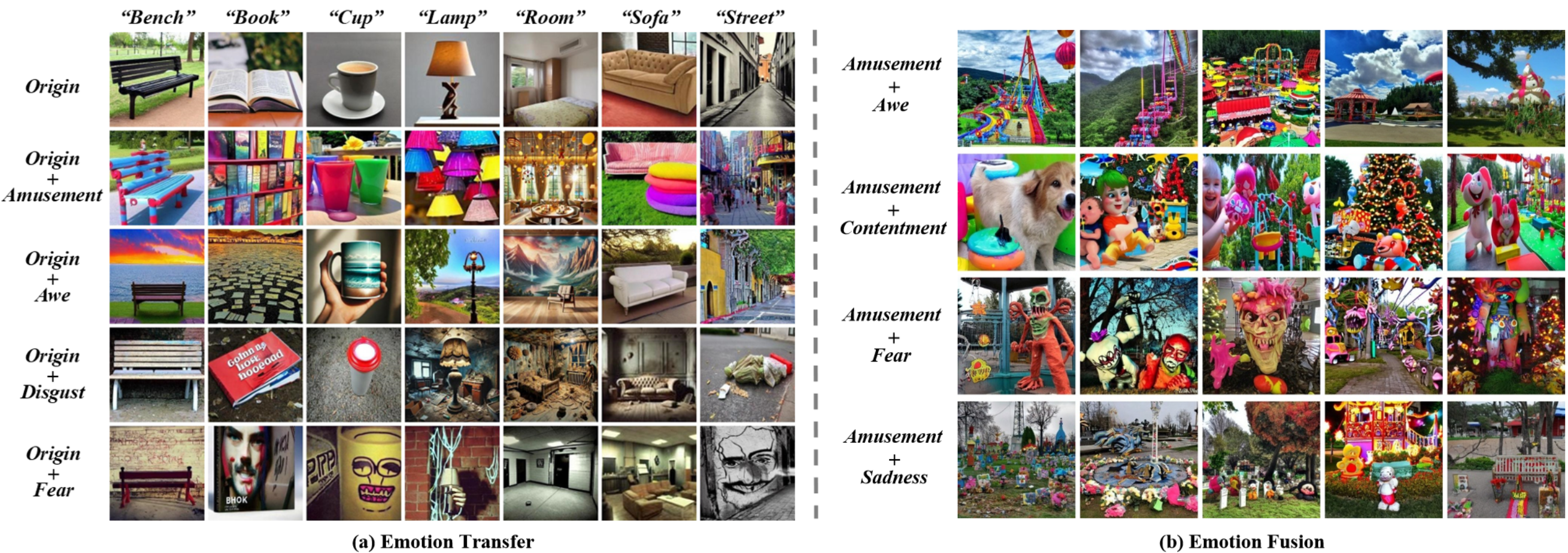}
            \vspace{-6pt}
            \caption{(a) Emotion transfer, where we combine each of the learned emotion representations with several neutral semantics. With such elements, one can automatically generate images with rich emotional content, enabling a nuanced and expressive visual synthesis. (b) Emotion fusion, where two distinct emotion representations are jointly integrated into the generative process. This allows the model to synthesize images that simultaneously convey both emotions, paving the way for more expressive and contextually rich visual emotion generation.}
            \label{fig:transfer_merged}
            \vspace{-13pt}
        \end{figure*}

 \begin{figure}
            \centering
            \includegraphics[width=0.95\linewidth]{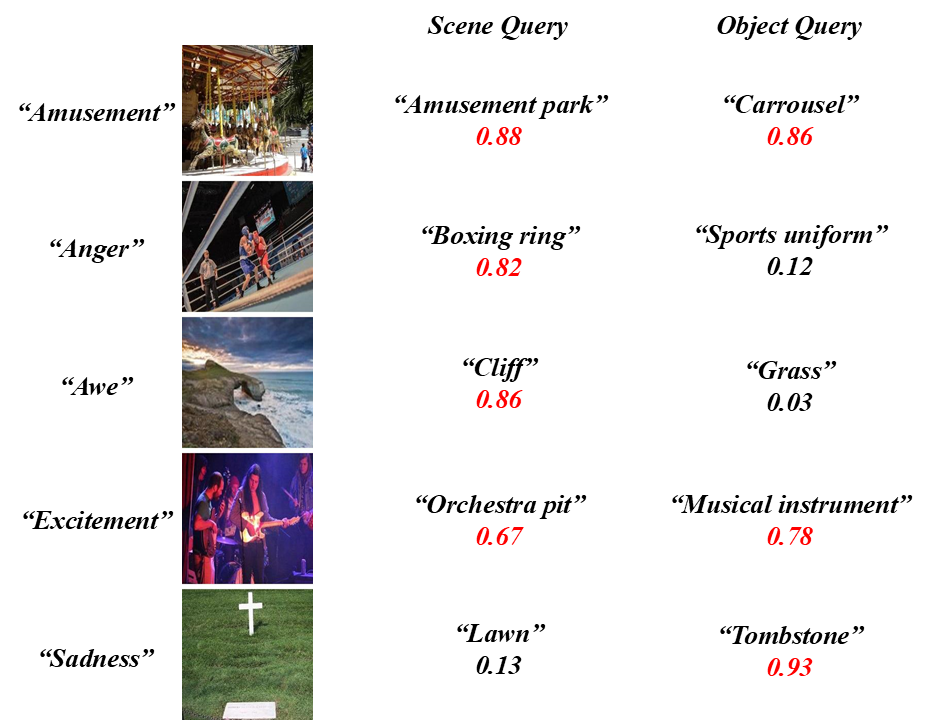}
            \vspace{-6pt}
             \caption{\textcolor{b}{Visualization of emotional correlation coefficients for expert queries. For each query, we show the most similar retrieved text attributes with their coefficients.}}
            \label{fig:coefficient_merged_v3}
            \vspace{-25pt}
        \end{figure}

\vspace{-16pt}
 \subsection{Extended Qualitative Analysis}
 \label{Qualitative}
\noindent{\textbf{Visualization of the Attention Maps for Expert Queries.}
 Fig.~\ref{fig:query_merged} presents a visualization of the attention maps for expert queries, demonstrating their ability to focus on distinct scene-level and object-level regions associated with various emotional labels. The \emph{Query} first attends to broader contextual areas, such as the concert stage in the \emph{Excitement} image and the natural environment in the \emph{Disgust} image, establishing an initial understanding of the scene that frames the emotional context. Subsequently, the \emph{Object Query} focuses on specific, emotion-evoking details within the images, such as the flames in the \emph{Anger} image and the bared teeth in the lion's open mouth in another \emph{Anger} image. This sequential attention mechanism, which progresses from scene-level to object-level extraction, highlights the effectiveness of our hierarchical emotional understanding chain, guided by expert queries.
 By first attending to the scene and then focusing on key objects, the model progressively refines its understanding of emotional features, supporting the inference of abstract emotional concepts. Leveraging both levels of information enables the model to interpret complex emotional cues and distinguish subtle variations across emotional categories.    

 \vspace{2pt}
 \noindent\textbf{Visualization of the Emotion Transfer.}
 We further combine emotions with neutral objects to create interesting and meaningful emotional creations, as shown in Fig.~\hyperref[fig:transfer_merged]{10(a)}. Each row illustrates the effect of transferring a distinct emotion onto a range of neutral objects, including \emph{Bench}, \emph{Book}, \emph{Cup}, \emph{Lamp}, \emph{Room}, \emph{Sofa}, and \emph{Street}. For example, applying \emph{Amusement} to \emph{Bench} results in a colorful, lively park bench surrounded by vibrant scenery, evoking a cheerful and playful atmosphere, while combining \emph{Amusement} with \emph{Lamp} creates a display of multi-colored, whimsical lighting fixtures. 
 These transformations highlight the model's flexibility in generating emotionally rich and contextually appropriate representations by integrating learned emotion features into neutral semantics. With such elements, one can automatically generate images with rich emotional content, enhancing visual storytelling and supporting applications where emotional impact is essential, such as digital art, advertising, and immersive media.

 \vspace{2pt}
 \noindent\textbf{Visualization of the Emotion Fusion.}
Moreover, to further demonstrate the expressive capacity of our framework, we investigate \emph{emotion fusion}, which synthesizes visual content conditioned on multiple emotional representations simultaneously rather than being restricted to a single label. As shown in Fig.~\hyperref[fig:transfer_merged]{10(b)}, such fusion generates complex and nuanced affective imagery that transcends simple emotion transfer. For example, combining \emph{Amusement} with \emph{Awe} produces vibrant and spectacular amusement park scenes, while \emph{Amusement + Contentment} yields warm and joyful imagery featuring playful pets and festive decorations. In contrast, fusions such as \emph{Amusement + Fear} introduce unsettling or eerie elements into otherwise cheerful contexts, and \emph{Amusement + Sadness} results in colorful amusement settings juxtaposed with somber atmospheres. 
Such capabilities are particularly valuable for applications that demand nuanced emotional storytelling, including digital art, film production, virtual reality, and advertising. 

 \vspace{2pt}
 \noindent\textbf{Visualization of the Emotional Correlation Coefficient for expert queries.}
\textcolor{b}{In Fig.~\ref{fig:coefficient_merged_v3}, we visualize the semantically closest text attributes associated with each expert query, showing that the queries capture both scene and object information. The \textquotedblleft Scene Query\textquotedblright{} identifies broader contextual settings, such as \textquotedblleft Amusement park\textquotedblright{} for \textquotedblleft Amusement\textquotedblright{} and \textquotedblleft Orchestra pit\textquotedblright{} for \textquotedblleft Excitement\textquotedblright{}, while the \textquotedblleft Object Query\textquotedblright{} focuses on salient entities, such as \textquotedblleft Carousel\textquotedblright{} and \textquotedblleft Musical instrument\textquotedblright{}. The visualization also highlights that different attributes have different emotional correlation coefficients: emotion-relevant attributes receive higher coefficients, whereas more neutral attributes, such as \textquotedblleft Grass\textquotedblright{} or \textquotedblleft Lawn\textquotedblright{}, have lower values. Using these coefficients in query fusion enables differentiated weights, improving emotion fidelity while preserving rich visual details in the generated images.}

\vspace{-12pt}
\subsection{Extended Qualitative Discussion}
\noindent{\textbf{Discussion on Computational Cost and Scalability.~}\textcolor{b}{
To assess practicality, we compare training and inference costs with representative baselines in Tab.~\ref{tab:efficiency_analysis}. For understanding, our method achieves 235.85 FPS (4.24 ms/image), outperforming EmoVIT (80.25 FPS). For generation, our latency (4.25 s/image) is comparable to EmoGen (4.21 s/image) while yielding better performance. Overall, our training time (14.2 h) is lower than both baselines (14.8--15.4 h), indicating a favorable trade-off. The overall pipeline scales well: training is naturally data-parallel and can benefit directly from additional GPUs.  The feedback stage is embarrassingly parallel, and its computation is controllable via the ratio of synthesized data (20\% as a balanced choice). Finally, UniEmo remains effective when scaling the diffusion backbone from SD~1.5 to SD~XL, suggesting good scalability to stronger generative models.}

\begin{table}[t]
    \footnotesize
    \caption{\textcolor{b}{Computational complexity comparison. EmoVIT is an understanding-only model, and EmoGen is a generation-only model. All timings are measured under the same hardware and identical evaluation settings to ensure fair comparison.}}
    \label{tab:efficiency_analysis}
    \vspace{-6pt}
    \centering
    
    \setlength{\tabcolsep}{2pt} 
    
        \begingroup
    \color{b} 
    \begin{tabularx}{\linewidth}{l >{\centering\arraybackslash}X >{\centering\arraybackslash}X >{\centering\arraybackslash}X}
        \toprule
        
        \multirow{2}{*}{\textbf{Method}} & 
        \multirow{2}{*}{\textbf{\begin{tabular}{@{}c@{}}Training\\Time (h)\end{tabular}}} & 
        \multicolumn{2}{c}{\textbf{Inference Time}} \\

        \cmidrule(lr){3-4}
        
        & & \textbf{Understanding} & \textbf{~~~Generation} \\
        \midrule
        
        EmoVIT (Understanding Only) & 
        15.4 & 
        \begin{tabular}{@{}c@{}}12.46 ms\\(80.25 FPS)\end{tabular} & 
        / \\
        
        EmoGen (Generation Only) & 
        14.8 & 
        / & 
        \textbf{4.21 s/img} \\
        \hline
        
        \textbf{Ours (Both)} & 
        \textbf{\begin{tabular}{@{}c@{}}14.2\\(Stage 1+2)\end{tabular}} & 
        \textbf{\begin{tabular}{@{}c@{}}4.24 ms\\(235.85 FPS)\end{tabular}} & 
        4.25 s/img \\
        \bottomrule
    \end{tabularx}
     \endgroup
    \vspace{-8pt}
\end{table}

\vspace{2pt}
\noindent{\textbf{Discussion on the Robustness of Attribute Annotations.}} 
\textcolor{b}{We evaluate robustness to attribute noise by randomly replacing emotion-related attributes with wrong category samples at 5\%, 10\%, and 30\% ratios (Tab.~\ref{tab:robustness_noise}). Performance remains largely stable under moderate noise (5--10\%), while a clear degradation at 30\% indicates that severe corruption can weaken the conditioning signal. We further test practicality by replacing manual attributes with automatically discovered ones from off-the-shelf classifiers, using an ImageNet-pretrained object classifier and a PLACES365-pretrained scene classifier. The performance remains comparable in Tab.~\ref{tab:robustness_automatic} with only minor changes, suggesting that UniEmo is not dependent on manual attribute annotations and is robust to the attribute source.}


\begin{table}[t]
    \footnotesize
   \caption{\textcolor{b}{Robustness to attribute noise. We randomly replace 5\%, 10\%, or 30\% of emotion-related attributes with attributes from wrong emotion categories to simulate noise.}}

     \vspace{-6pt}
    \label{tab:robustness_noise}
    \centering
    \setlength{\tabcolsep}{2.5pt} 
    \begingroup
    \color{b}
        \begin{tabular}{l c c c c c c}
            \toprule
            \textbf{Source} & 
            \textbf{Accuracy} $\uparrow$ & 
            \textbf{FID} $\downarrow$ & 
            \textbf{LPIPS} $\uparrow$ & 
            \textbf{Emo-A} $\uparrow$ & 
            \textbf{Sem-C} $\uparrow$ & 
            \textbf{Sem-D} $\uparrow$ \\
            \midrule
            
            Baseline &85.30\% &
            27.73 & 0.793 & 79.66\% & 0.640 & 0.0383 \\
            
            5\% &85.24\% &
            27.81 & 0.786 & 79.43\% & 0.638 & 0.0381 \\
            
            10\% &84.62\% &
            28.24 & 0.781 & 78.71\% & 0.624 & 0.0367 \\
            
            30\% &82.23\% &
            34.59 & 0.723 & 75.05\% & 0.529 & 0.0303 \\
            \bottomrule
        \end{tabular}
    \endgroup
    \vspace{-15pt}
\end{table}
\begin{table}[t]
    \footnotesize
\caption{\textcolor{b}{Robustness to automatic attributes. We replace manual attributes with off-the-shelf discovered ones.}}

     \vspace{-6pt}
    \label{tab:robustness_automatic}
    \centering
    \setlength{\tabcolsep}{2.5pt} 
    \begingroup
    \color{b}
        \begin{tabular}{l c c c c c c}
            \toprule
            \textbf{Noise} & 
            \textbf{Accuracy} $\uparrow$ & 
            \textbf{FID} $\downarrow$ & 
            \textbf{LPIPS} $\uparrow$ & 
            \textbf{Emo-A} $\uparrow$ & 
            \textbf{Sem-C} $\uparrow$ & 
            \textbf{Sem-D} $\uparrow$ \\
            \midrule
            
            Manual &85.30\% &
            27.73 & 0.793 & 79.66\% & 0.640 & 0.0383 \\
            Automatic &85.16\% &
            27.94 & 0.784 & 79.31\% & 0.633 & 0.0371 \\
            \bottomrule
        \end{tabular}
    \endgroup
    \vspace{-6pt}
\end{table}

   \begin{table}
              \caption{\textcolor{b}{Robustness of the emotional correlation coefficient $\alpha$.
We evaluate $\alpha$ under three settings: (i) \textit{Classifier Bias}, where we perturb the pretrained classifier by adding Gaussian noise to its logits for a fraction of samples (\(\rho{=}5\%,10\%,30\%\)); (ii) \textit{Cross-Dataset} transfer (FI$\rightarrow$EmoSet), where $\alpha$ computed on FI is directly applied to EmoSet; and (iii) \textit{Long-Tail}, where $\alpha$ is estimated on a class-imbalanced split.}}

        \label{tab:robustness_Correlation}
        \vspace{-6pt}
        \centering
        \renewcommand\arraystretch{1.18}
        \footnotesize
        \setlength{\tabcolsep}{3pt}
        \begingroup
        \color{b}
        \begin{tabular}{lccccccccc}
            \toprule
            \textbf{Settings} & \textbf{Accuracy} $\uparrow$&\textbf{FID $\downarrow$} & \textbf{LPIPS $\uparrow$} & \textbf{Emo-A $\uparrow$} & \textbf{Sem-C $\uparrow$} & \textbf{Sem-D $\uparrow$}  \\
		  \hline
            \rowcolor{gray!20} 
            \multicolumn{7}{c}{\textit{Classifier Bias}} \\
        \hline
Baseline  &85.30\% &
            27.73 & 0.793 & 79.66\% & 0.640 & 0.0383 \\
$\rho{=}5\%$&85.13\%   &27.92 & 0.777 &79.31\%  &0.633  &0.0375      \\
$\rho{=}10\%$  &84.65\%  &28.53  &0.775  &78.94\%  &0.621  &0.0364  \\
$\rho{=}30\%$  &81.76\%  &35.67  &0.703  &73.43\%  &0.535  &0.0304  \\
            \hline
            \rowcolor{gray!20} 
            \multicolumn{7}{c}{\textit{Cross DataSet (FI$\rightarrow$EmoSet)}} \\
            \hline
             Baseline&85.30\% &
            27.73 & 0.793 & 79.66\% & 0.640 & 0.0383 \\
              Transfer&85.04\% &28.01&0.780	&79.22\%	&0.635	&0.0373	 \\
            \hline
            \rowcolor{gray!20} 
            \multicolumn{7}{c}{\textit{Long-Tail}} \\
            \hline
           Baseline&85.30\% &
            27.73 & 0.793 & 79.66\% & 0.640 & 0.0383 \\
              Long-Tail&84.99\% &28.13&0.774	&79.07\%	&0.631	&0.0370	 \\
            \bottomrule
        \end{tabular}
        \endgroup
        \vspace{-6pt}
    \end{table}
\begin{table}[!t]
    \footnotesize
    \caption{\textcolor{b}{Performance on the multi-emotion benchmarks.}}
    \vspace{-6pt}
    \label{tab:multi_emotion_perf}
    \centering
    \setlength{\tabcolsep}{4.5pt}
    \begingroup
    \color{b}
    \begin{tabular}{l c c c c}
        \toprule
        \textbf{Method} & 
        \textbf{Hamming} $\downarrow$ & 
        \textbf{Ranking} $\downarrow$ & 
        \textbf{Micro-F1} $\uparrow$ & 
        \textbf{Macro-F1} $\uparrow$ \\
        \hline
              \rowcolor{gray!20} 
            \multicolumn{5}{c}{\textit{Emotic Benchmark}} \\
            \hline
        EmoVIT & 
        0.131 & 
        0.146 & 
        0.203 & 
        0.049 \\
             UniEmo (Ours) & 
       0.115& 0.125
         &0.214 
         &0.063 
         \\
          \hline
              \rowcolor{gray!20} 
            \multicolumn{5}{c}{\textit{Emotion6 Benchmark}} \\
            \hline
                    EmoVIT & 0.144
         & 0.173
         & 0.811
        & 0.782
         \\
             UniEmo (Ours) &0.107 
         & 0.154
         & 0.876
         & 0.840
         \\
            \hline
    \end{tabular}
    \endgroup
    \vspace{-12pt}
\end{table}
    \vspace{2pt}
\noindent{\textbf{Discussion on the Robustness of Emotional Correlation Coefficient.}} 
\textcolor{b}{We evaluate the robustness of the emotional correlation coefficient \(\alpha\) under three settings (Tab.~\ref{tab:robustness_Correlation}). 
\textbf{\textit{(i) Classifier Bias:}} during \(\alpha\) estimation, we add Gaussian noise to the pretrained classifier logits for a fraction \(\rho\in\{5\%,10\%,30\%\}\) of samples. Performance is stable under moderate noise (\(\rho\leq10\%\); e.g., Acc. \(85.30\%\rightarrow84.65\%\), FID \(27.73\rightarrow28.53\)), but drops at \(\rho{=}30\%\), showing robustness to realistic misclassification while extreme bias can degrade the signal. 
\textbf{\textit{(ii) Cross-dataset:}} we directly transfer \(\alpha\) computed on FI to EmoSet (FI\(\rightarrow\)EmoSet) without recomputation, and observe only minor differences from the EmoSet-estimated baseline, suggesting that the coefficient remains largely reliable under dataset shift.
\textbf{\textit{(iii) Long-tail:}} we estimate \(\alpha\) on a class-imbalanced (long-tail) split constructed by class-wise down-sampling. Concretely, we keep the full data for head emotion categories and randomly down-sample the tail categories to a much smaller fixed fraction, yielding a pronounced head-to-tail imbalance, while keeping the evaluation protocol unchanged. Under this long-tail setting, performance shows only slight degradation, indicating that \(\alpha\) remains reasonably stable even when emotion frequencies are highly imbalanced.
Overall, \(\alpha\) serves as a robust dataset-level prior under moderate classifier noise and distribution shifts, while extreme perturbations can weaken conditioning.}

\vspace{2pt}
\noindent{\textbf{Discussion on Performance on Multi-Emotion Benchmarks.}}
\textcolor{b}{We qualitatively demonstrate UniEmo's emotion fusion ability in Fig.~\hyperref[fig:transfer_merged]{10(b)}. Following prior work~\cite{feng2023probing}, we further evaluate UniEmo quantitatively on multi-label emotion benchmarks, including Emotic~\cite{kosti2017emotion} and Emotion6~\cite{peng2015mixed}. 
Concretely, we adapt the understanding branch to multi-label prediction by replacing the softmax classifier with a sigmoid-based head and optimizing with binary cross-entropy, so that each image can be assigned multiple emotions. We report standard multi-label metrics, including Hamming loss, ranking loss, and Micro/Macro-F1. As shown in Tab.~\ref{tab:multi_emotion_perf}, UniEmo consistently outperforms EmoVIT on both datasets, achieving lower Hamming and ranking losses and higher Micro/Macro-F1 scores. These results indicate that UniEmo can natively support multi-emotion recognition.}

    
        

        
        
        
    

\section{CONCLUSION}
In this paper, we introduce UniEmo, a unified framework that integrates emotional understanding with generation. By incorporating learnable expert queries, UniEmo transforms emotional understanding into a coarse-to-fine visual processing chain and conditions a diffusion model to generate emotion-evoking images. To enhance the diversity and fidelity of the generated emotional images, we introduce the emotional correlation coefficient and emotional condition loss into the fusion process. Ultimately, we leveraged UniEmo's generation capabilities to filter high-quality emotional images, further improving performance. 
Extensive experiments show that UniEmo significantly outperforms state-of-the-art methods in both emotional understanding and generation tasks.
\ifCLASSOPTIONcaptionsoff
  \newpage
\fi

\bibliographystyle{IEEEtran}
\bibliography{IEEEabrv,reference}

\end{document}